\documentclass[10pt,twocolumn,letterpaper]{article}

\usepackage{nameref}
\usepackage{varioref}
\usepackage[pagenumbers]{cvpr} 
\usepackage{multirow}
\usepackage{amsmath}

\definecolor{cvprblue}{rgb}{0.21,0.49,0.74}
\usepackage[pagebackref,breaklinks,colorlinks,allcolors=cvprblue]{hyperref}

\def\paperID{5554} 
\def\confName{CVPR}
\def\confYear{2025}

\title{ModeDreamer: Mode Guiding Score Distillation for\\ Text-to-3D Generation using Reference Image Prompts}

\author{%
 Uy Dieu Tran$^{1}$\thanks{First two authors contribute equally.} \quad  Minh Luu$^{1}$\footnotemark[1] \\ 
 Phong Nguyen$^{1}$ \quad Khoi Nguyen$^{1}$ \quad Binh-Son Hua$^{1, 2}$ \\ \\
  $^1$VinAI Research, Vietnam \quad
  $^2$Trinity College Dublin, Ireland
 \\
\tt\small \{v.uytd,v.minhlnh,v.phongnh31,v.khoindm\}@vinai.io \qquad binhson.hua@tcd.ie \\
}

\def\1n{\mathbf{1}_n}
\def\0{\mathbf{0}}
\def\1{\mathbf{1}}

\definecolor{pink}{rgb}{0.9,0.5,0.5}
\definecolor{purple}{rgb}{0.5, 0.4, 0.8}   
\definecolor{gray}{rgb}{0.3, 0.3, 0.3}
\definecolor{mygreen}{rgb}{0.2, 0.6, 0.2}

\definecolor{greena}{rgb}{0.4, 0.5, 0.1}

\definecolor{bluea}{rgb}{0, 0.4, 0.6}

\definecolor{reda}{rgb}{0.6, 0.2, 0.1}

\newcommand{\cm}[1]{}

\newcommand{\myheading}[1]{\vspace{1ex}\noindent \textbf{#1}}

\newif\ifshowsolution
\showsolutiontrue

\ifshowsolution  

\else

\fi

\definecolor{gray}{rgb}{0.3, 0.3, 0.3}

\begin{document}
\maketitle
\begin{abstract}
Existing Score Distillation Sampling (SDS)-based methods have driven significant progress in text-to-3D generation. However, 3D models produced by SDS-based methods tend to exhibit over-smoothing and low-quality outputs. These issues arise from the mode-seeking behavior of current methods, where the scores used to update the model oscillate between multiple modes, resulting in unstable optimization and diminished output quality. To address this problem, we introduce a novel image prompt score distillation loss named ISD, which employs a reference image to direct text-to-3D optimization toward a specific mode. Our ISD loss can be implemented by using IP-Adapter, a lightweight adapter for integrating image prompt capability to a text-to-image diffusion model, as a mode-selection module. A variant of this adapter, when not being prompted by a reference image, can serve as an efficient control variate to reduce variance in score estimates, thereby enhancing both output quality and optimization stability. Our experiments demonstrate that the ISD loss consistently achieves visually coherent, high-quality outputs and improves optimization speed compared to prior text-to-3D methods, as demonstrated through both qualitative and quantitative evaluations on the T3Bench benchmark suite.
\end{abstract}    
\section{Introduction}
\begin{figure*}[!htp]
  \centering
  \includegraphics[width=0.99\linewidth]{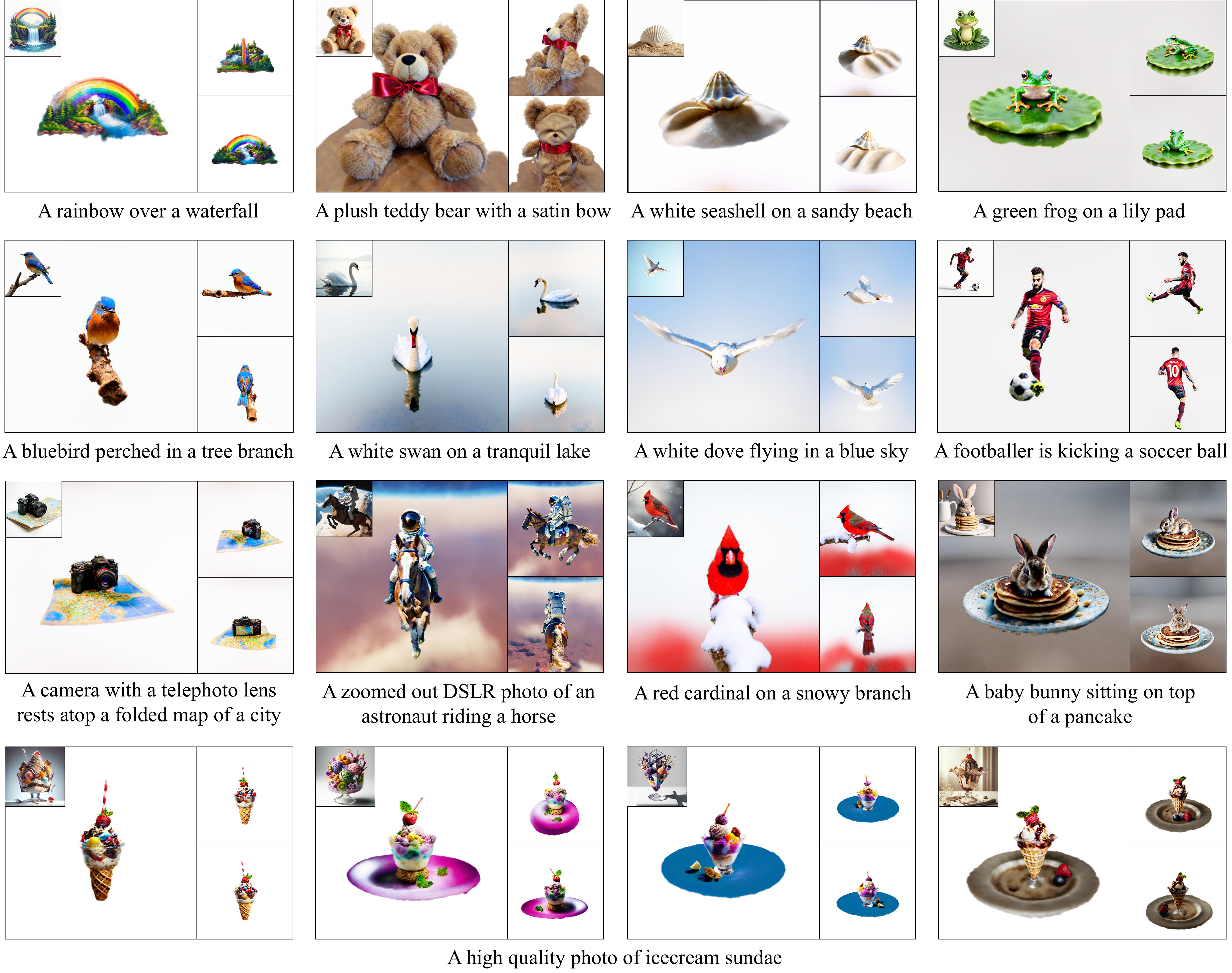}
   \caption{Our mode-guiding score distillation using the ISD loss explicitly selects a desired mode captured in a diffusion prior using a reference image prompt to steer text-to-3D generation. Our method leads to high-quality and diverse 3D generation. For each sub-figure, the top-left illustrates the reference image used for mode selection, and the remaining images illustrate different views of the generated object. The last row demonstrates the diversity of our 3D generation by using  different reference images on the same text prompt.}
   \label{fig:teaser}
\end{figure*}

The generation of 3D assets continues to be a dynamic and influential research field within computer vision and computer graphics, driven by diverse applications in gaming, film, e-commerce, and robotics. High-quality 3D content creation is resource-intensive, making automated 3D generation a key research objective. Recent advancements in neural scene representations~\cite{mildenhall2020nerf,kerbl3Dgaussians} and diffusion-based generative models~\cite{stable-diffusion,esser2024scaling} have led to significant progress in producing creative 3D content from text prompts~\cite{poole2023dreamfusion, wang2023prolificdreamer}.

To generate a 3D model from a text prompt, a notable approach is to optimize a 3D representation using distillation-based method, which utilizes a pretrained text-to-image diffusion model to guide the generation of 3D objects. Since DreamFusion~\cite{poole2023dreamfusion}, SDS-based methods~\cite{wang2023prolificdreamer,qiu2024richdreamer,Ma2023GeoDream,qiu2024richdreamer} offer superior 3D generation quality, albeit being more expensive in its optimization to generate 3D models. These methods have been designed to address the limitations of the original SDS, mitigating over-smoothing and over-saturation effects.

At its core, score distillation sampling aims to optimize the KL-divergence between the Gaussian distributions in a forward diffusion process and the distribution of a pretrained diffusion model~\cite{poole2023dreamfusion}. 
One challenge in this optimization process is the mode-seeking behavior of the KL-divergence, which is difficult to control and has limited capability to produce diverse and high-quality 3D objects.
Some previous methods attempted to address the limitations of SDS loss by employing variational score distillation (VSD) \cite{wang2023prolificdreamer}, reducing gradient variance \cite{liu2016stein}, or utilizing two-stage training strategies \cite{magic3d,Metzer_2023_CVPR, Yang2024VP3D}. Furthermore, according to the leaderboards from T3bench \cite{t3bench} and GPTEval3D \cite{wu2023gpteval3d}, the quality of 3D objects generated by VSD remains superior, despite significantly slow generation time.

In this paper, we address such mode-seeking issues by directly guiding mode convergence using image prompts during score distillation sampling. 
Our idea is to view the text-to-image diffusion prior as an integral of conditional distributions conditioned on reference image prompts. 
This conceptual change allows us to use a reference image to directly guide score distillation sampling to converge to a favorable mode specified in the reference, thereby resolving the ambiguity faced in mode seeking.
We design our method based on IP-Adapter~\cite{ipadapter}, a lightweight adapter for pretrained text-to-image diffusion models to integrate guidance from image prompts.
By exploiting the positive correlation between the noise prediction by the diffusion prior with and without a reference image prompt, respectively, we devise \emph{Image prompt Score Distillation (ISD)}, a novel distillation with lower variance in its score estimates due to a more effective control variate based on our adapter. 
This strategy enables us to effectively steer and control the score distillation process, resulting in diverse high-quality 3D outputs that capture the variability and fidelity of the reference images. 
To mitigate the multi-face Janus artifact, we combine the advantages of our proposed ISD loss with an enhanced SDS variant that incorporates a multi-view diffusion prior, such as MVDream~\cite{shi2024mvdream} or Zero123~\cite{zero123,shi2023zero123plus}. This integration promotes cross-view consistency and significantly enhances the overall quality of the 3D outputs.

We evaluate our method on the recently introduced T3Bench~\cite{he2023t3bench}, a text-to-3D benchmark containing diverse text prompts across three complexity levels, specifically designed for the 3D generation task, and compare it with state-of-the-art text-to-3D score distillation methods. Both qualitative and quantitative results demonstrate the effectiveness of our approach over previous methods. Our method produces 3D assets with realistic appearance and shape, and is capable of generating a diverse range of 3D objects from the same text prompt, as demonstrated in Fig.~\ref{fig:teaser}.
In summary, our main contributions are as follows:

\begin{itemize} 
    \item We analyze the mode-seeking behavior of the SDS loss and propose to leverage the visual prompting capabilities of IP-Adapter~\cite{ipadapter} to effectively guide and stabilize the noisy score distillation for text-to-3D generation.
    
    \item We introduce a novel variate control term in score distillation, which reduces variance in score estimates. We achieve high-quality 3D generation without the need of LoRA training in an alternative optimization fashion as done in previous methods like VSD~\cite{wang2023prolificdreamer}.
    
    \item Our method achieves high-quality and diverse 3D results comparable to state-of-the-art approaches~\cite{qiu2024richdreamer,Ma2023GeoDream,ma2024scaledreamer}, with the added advantage of significantly reduced optimization time, requiring only 30-40 minutes compared to the several hours of training needed by VSD~\cite{wang2023prolificdreamer}.
\end{itemize}

\section{Related Work}
\label{sec:related_work}
\myheading{Text-to-Image Generation.} 
Early text-to-image generation extended GAN architectures \cite{goodfellow2014generative} to condition on textual inputs by mapping text embeddings into the image space, but these faced challenges with training stability and limited diversity due to mode collapse. Diffusion models~\cite{stable-diffusion, dall-e, saharia2022photorealistic, yu2022scaling, anonymous2024eliminating} significantly advanced the field by leveraging large image-text pair datasets for modeling the relationships between text and images distribution through iterative denoising processes, offering higher image quality, diversity, and improved stability over GANs. IP-Adapter~\cite{ipadapter} further enhanced alignment with user intent through a decoupled cross-attention mechanism incorporating both text and image inputs. Our approach builds on IP-Adapter by additionally conditioning on a reference image.

\myheading{Image/Text-to-Multiview Synthesis.} Consistent multi-view image generation is crucial for many applications. Several methods ~\cite{zero123, shi2023zero123plus, wang2023imagedream} can generate novel views from an input image by fine-tuning pretrained text-to-image models on large-scale 3D datasets~\cite{objaverse, objaverseXL}, accounting for camera positions. These methods model view-dependent effects and occlusions to ensure coherence across multiple viewpoints. Text-to-multiview methods like MVDream~\cite{shi2024mvdream} extend text-to-image generation to multi-view synthesis by fine-tuning pretrained Stable Diffusion with a 3D-aware generative module on 3D data. Our method leverages multi-view prior of MVDream to mitigate the Janus problem.

\myheading{Image-to-3D Generation.} Recent approaches reconstruct 3D models from a single input image using pre-trained diffusion models \cite{hong2024lrm, gslrm2024, TripoSR2024, realfusion, liu2023one2345, liu2024syncdreamer, tang2024lgm, consistent123, shen2024gamba, wang2025crm, xu2024instantmesh, li2024craftsman}. These models can generate 3D objects within seconds from real or synthesized images without per-prompt optimization, significantly reducing computational demands compared to traditional techniques. Despite their impressive results, they struggle with complex scenes involving intricate geometries or textures, especially thin structures, leading to reduced detail or incomplete reconstructions. Additionally, they heavily rely on large-scale 3D datasets \cite{objaverseXL,objaverse} for training, but since these datasets are smaller than image-text datasets like LAION-5B \cite{schuhmann2022laion5bopenlargescaledataset}, the models have limited generalization to unseen objects and struggle with images containing multiple objects or complex lighting conditions and backgrounds.

\begin{figure}[t]
  \centering
  \includegraphics[trim={0, 0.2cm, 0, 0.2cm}, clip,width=\linewidth]{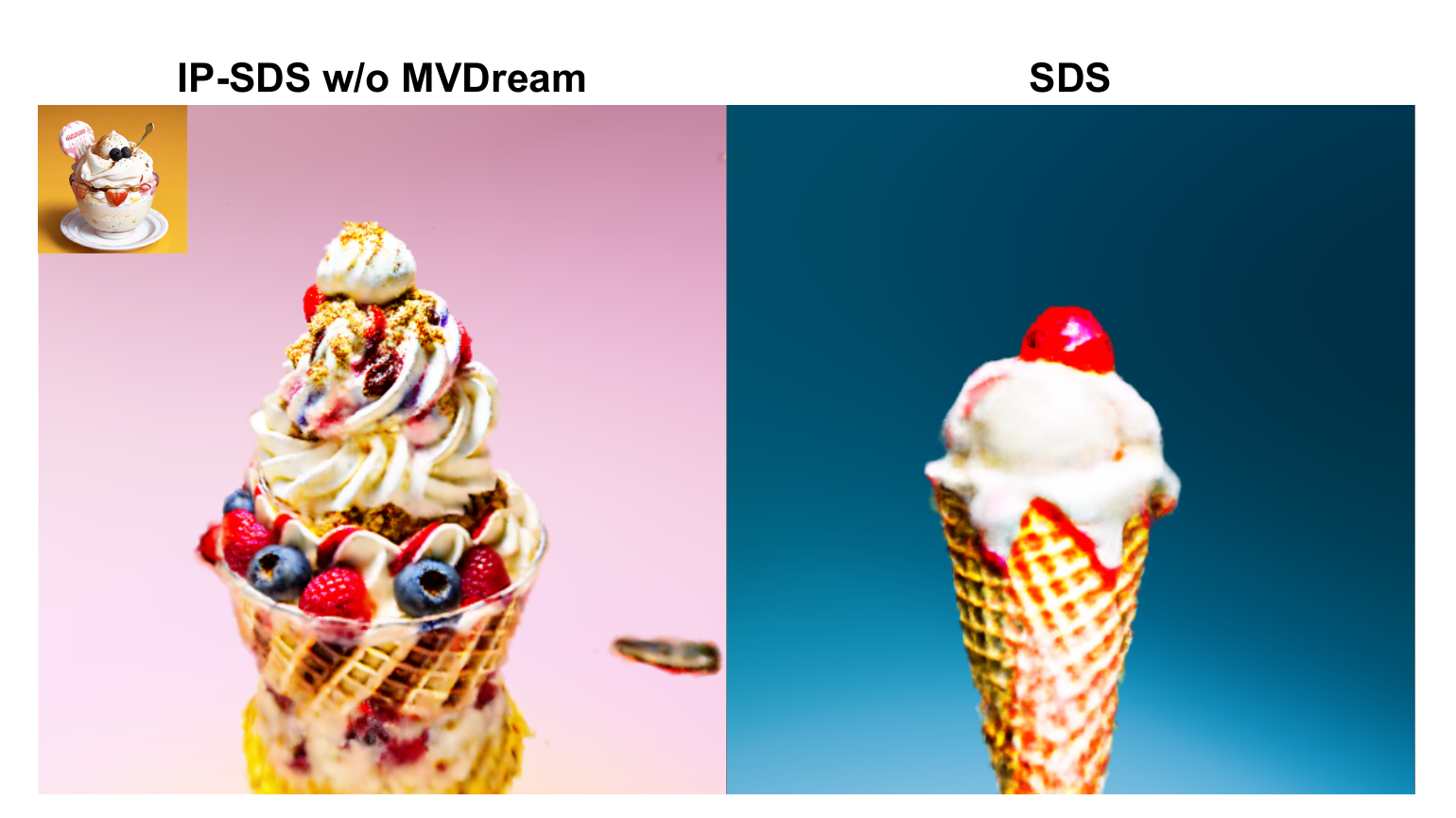}
   \caption{Comparison between vanilla SDS and IP-SDS, IP-SDS can generate detail and sharp texture while original SDS cannot.}
   \label{fig:sds_3d}
\end{figure}

\begin{figure}[t]
  \centering
  \includegraphics[trim={0, 1.2cm, 0, 0.2cm}, clip, width=\linewidth]{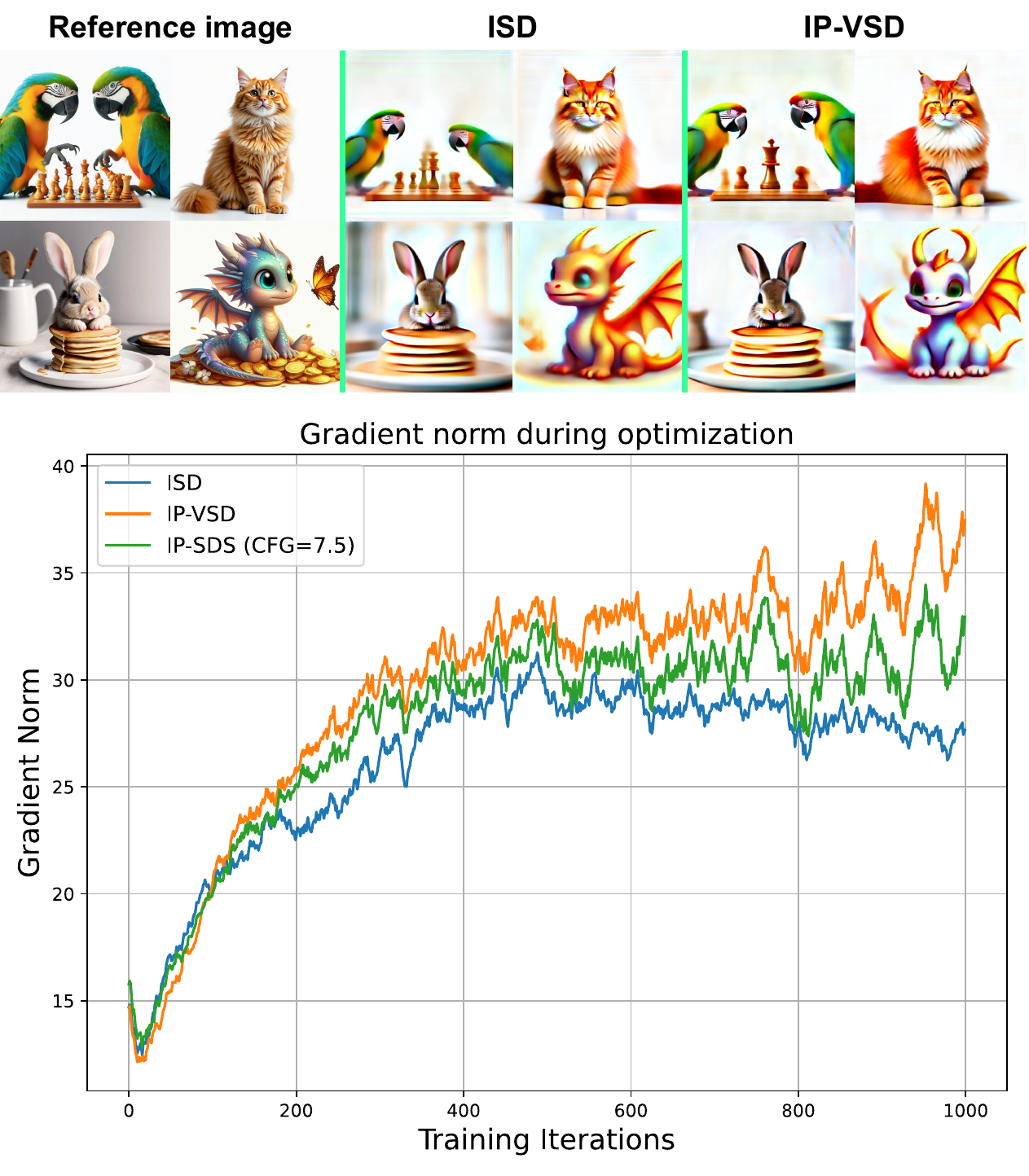}
   \caption{A 2D toy experiment. Our ISD can generate results similar to IP-VSD while having lower gradient variance. This advocates designing better control variates, discarding the need of training LoRA in an alternative fashion like in VSD.}
   \label{fig:gradient grad}
\end{figure}

\myheading{Text-to-3D Generation.} Training-based methods~\cite{tang2024lgm, li2024instantd, qiu2024richdreamer, xie2024latte3d} learn a direct mapping from text to 3D representations through supervised learning on large 3D datasets~\cite{objaverse, objaverseXL}, but require extensive annotated data and significant computational resources. In contrast, optimization-based methods iteratively optimize a 3D representation to align with a text prompt, guided by pretrained text-to-image models. Foundational approaches~\cite{dreamfields, poole2023dreamfusion} use pretrained diffusion models to guide the optimization of neural radiance fields (NeRFs)\cite{mildenhall2020nerf} by minimizing the difference between rendered images and the target distribution conditioned on the prompt.
However, generating a 3D model for each prompt is resource-intensive. To address this, Lorraine et al.~\cite{lorraine2023att3d} propose an amortized optimization approach by training a unified model on multiple prompts simultaneously. This approach shares computation across prompts, reducing training time compared to per-prompt optimization. 
Other approaches focus on improving the visual quality and fidelity of generated 3D models~\cite{wang2023prolificdreamer, yu2024texttod, katzir2024noisefree, fantasia3d, huang2024dreamtime, lee2024dreamflow, Yang2024VP3D, Ma2023GeoDream, qiu2024richdreamer, ma2024scaledreamer, jiang2025jointdreamer, ding2024text} or computational intensity ~\cite{tang2024dreamgaussian, yi2023gaussiandreamer} by utilizing more efficient 3D representations, such as Gaussian Splatting~\cite{kerbl3Dgaussians}. Some tackle mode collapse in SDS loss~\cite{tran2025diverse, yan2024flow}, enabling diverse 3D content generation from the same text. For example, DiverseDream~\cite{tran2025diverse} encourages the exploration of multiple plausible solutions to generate diverse outputs from a single text prompt. VP3D~\cite{Yang2024VP3D} employs an image-conditioned diffusion model to supervise the creation of 3D models, with multiview regularization being novel views synthesized from Zero123 \cite{zero123}. This approach may lead to inconsistent 3D results due to two primary factors: 1) the inherent inconsistencies of Zero123 \cite{zero123} and 2) the underperformance of the image-conditioned diffusion model in novel view conditioning. Our approach, using ISD loss, further mitigates the mode-seeking behavior of the SDS by leveraging guidance from reference images while ensuring consistent views of 3D objects.

\section{Background and Analysis}
\label{sec:background}

To generate coherent 3D objects without direct 3D data, the priors of a Latent Diffusion Model (LDM) can be leveraged to guide the creation of standard 3D representations, such as NeRF \cite{mildenhall2020nerf} or 3D Gaussian splatting \cite{kerbl3Dgaussians}. Specifically, a 3D representation parameterized by 
$\theta$ is iteratively optimized to align its rendered images 
$x=g(\theta, c)$ with LDM's distribution given a text prompt 
$y$.

\subsection{Score distillation sampling (SDS)}  
DreamFusion \cite{poole2023dreamfusion} proposed SDS loss to generate 3D models. The loss is formulated as:
\begin{align} \label{eq:sds_loss_form}
    \text{KL}(q(z_t|x=g({\theta}, c) \| p_{\phi}(z_t|y)),
\end{align}
where $q(z_t|x=g({\theta}, c))$ is a Gaussian distribution representing a forward diffusion process of the rendered images, and $p_{\phi}(z_t|y)$ denoting the marginal distribution at timestep $t$ of a pretrained LDM model. The gradient is calculated as:
\begin{align} \label{eq:sds}
    \nabla_\theta \mathcal{L}_{\text{SDS}} \triangleq \mathbb{E}_{t, \epsilon,c} \left[\omega(t)(\epsilon_{\phi} (x_t, t, y)-\epsilon) \frac{\partial g(\theta, c)}{\partial \theta}\right],
\end{align}
where $\omega(t)$ is a time-dependent weighting function, $\epsilon_{\phi}(.)$ is the predicted noise of LDM given the noisy input image $x_t=\alpha_t x + \sigma_t\epsilon$ created by adding Gaussian noise $\epsilon$ to the rendered image $x$ at timestep $t$ with noise scheduling coefficients $\alpha_t, \sigma_t$. 
Derived from the reverse KL divergence in \cref{eq:sds_loss_form}, the SDS loss exhibits mode-seeking behavior, aligning a single-mode Gaussian distribution with the complex, multimodal distribution of a pretrained LDM. This leads to oversaturated, oversmoothed, and low-diversity 3D models, as empirically analyzed in~\cite{poole2023dreamfusion}.

\begin{figure*}[!htp]
  \centering
  \includegraphics[trim={0, 0.0cm, 0, 0.0cm}, clip,width=1.\linewidth]{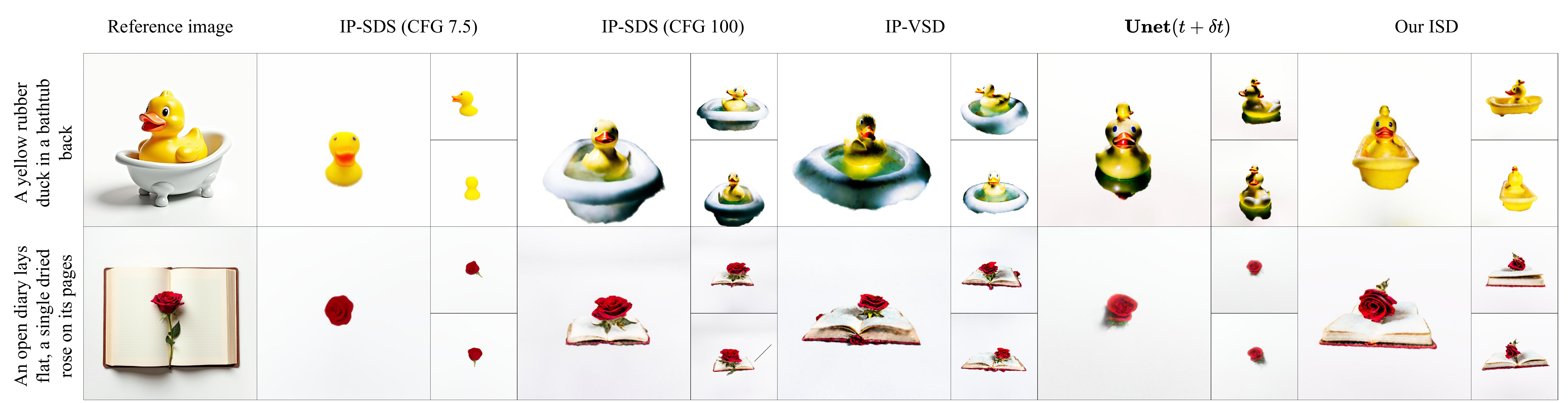}
   \caption{Different control variate settings including Gaussian noise $\epsilon$, learning LoRA-Unet in VSD \cite{wang2023prolificdreamer}, Unet at different timestep in ScaleDreamer \cite{ma2024scaledreamer}, and our ISD. Our method can generate 3D objects on par with VSD without learning additional LoRA-Unet.}
   \label{fig:control_variate}
\end{figure*}

Several enhancements have been introduced to address challenges with the SDS loss. ProlificDreamer~\cite{wang2023prolificdreamer} improves output quality by replacing the noise term in SDS with adaptive, trainable LoRA~\cite{hu2021lora} layers, which more effectively approximate 3D model scores. Although ProlificDreamer produces high-quality 3D results, it suffers from slow training speeds and instability due to the sensitivity of the LoRA layers.

Other methods, such as ScaleDreamer~\cite{ma2024scaledreamer} and CSD~\cite{CSD}, aim to accelerate training by approximating LoRA \cite{hu2021lora} layers using U-Net architectures with varying timesteps or by using only the CFG term in the distillation gradient. While they improve training efficiency, their 3D outputs often show oversaturation. Multi-stage strategies like Magic3D~\cite{magic3d}, Fantasia3D~\cite{fantasia3d}, and Latent NeRF~\cite{Metzer_2023_CVPR} enhance SDS by first training coarse geometry before optimizing texture. However, these methods still encounter issues with over-smoothness, over-saturation, and limited diversity due to reliance on SDS loss.
DiverseDream~\cite{tran2025diverse} improves 3D output diversity using textual inversion from reference images. However, it struggles with asymmetric object generation and the Janus problem due to bias toward reference views in visual prompts.

\subsection{Mode analysis}
\label{sec:method_analysis}

We hypothesize that the limitations of SDS loss come from oscillations in the modes. As the model samples noise differently at each iteration, the score updates may direct the model toward various modes within the data distribution. Consequently, the resulting 3D models may fail to converge to a single, stable mode. We speculate that if a high-quality mode can be identified in advance, the SDS loss is capable of generating high-quality samples. More specifically, we can express the marginal distribution of the LDM prior
$p_{\phi}(z_t|y)$ by integrating over reference images:
\begin{align} \label{eq:integral_mode}
    p_{\phi}(z_t|y) = \int_{x_{\text{ref}}}p_{\phi}(z_t|y,x_{\text{ref}})p(x_{\text{ref}}|y) \,dx_{\text{ref}},
\end{align}
where $x_{\text{ref}}$ denotes the reference image, $p_{\phi}(z_t|y,x_{\text{ref}})$ represents the LDM prior conditioned on the additional reference image along with text prompt $y$, and $p(x_{\text{ref}}|y)$ stands for the distribution of the reference image given a text prompt $y$. The \cref{eq:integral_mode} suggests that $p_{\phi}(z_t|y)$ is a combination of multiple modes, which complicates the optimization of the SDS loss. Since $x_{\text{ref}}$ provides concrete visual information, the model is less likely to oscillate between modes. Instead, it focuses on refining its output around the mode suggested by $x_{\text{ref}}$, i.e., $p_{\phi}(z_t|y,x_{\text{ref}})$. This significantly enhances stability during training and inference, as the model has a clearer target to optimize. Consequently, rather than randomly fluctuating among different interpretations of the prompt, the model can effectively concentrate its learning and sampling on the characteristics provided by $x_{\text{ref}}$.

\section{Our Method}
\label{sec:method}

\begin{figure*}[t]
  \centering
    \includegraphics[width=0.9\textwidth]{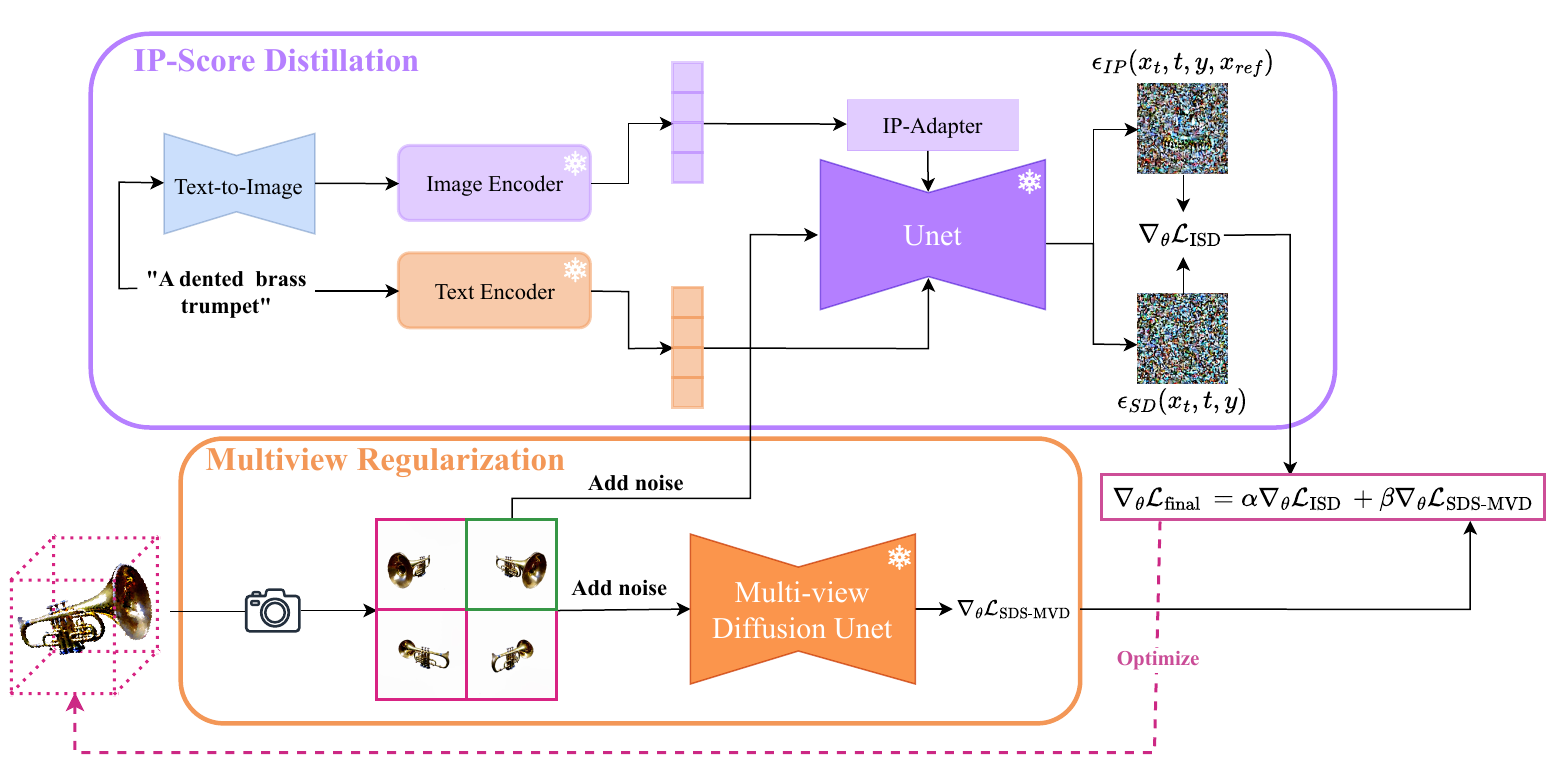}
   \caption{An overview of our method. Starting with input prompt $y$, we generate a reference image $x_{\text{ref}}$ using a text-to-image model. Both the text prompt and the image prompt are used with the IP-Adapter for score distillation, following our ISD gradient $\nabla_\theta \mathcal{L}_{\text{ISD}}$. To mitigate view bias by reference image and the Janus problem, we incorporate additional multi-view regularization by jointly optimizing $\nabla_\theta \mathcal{L}_{\text{ISD}}$ with $\nabla_\theta \mathcal{L}_{\text{SDS-MVD}}$. 
   }
   \label{fig:pipeline}
   \vspace{-4pt}
\end{figure*}

Based on the insight gained from mode analysis, we detail our method as follows. 
Starting with input prompt $y$, we generate a reference image $x_{\text{ref}}$ using a pretrained text-to-image diffusion model. 
We then calculate the score $\nabla_\theta \mathcal{L}_{\text{ISD}}$ that utilizes both the text prompt and the image prompt for 3D optimization. 
To mitigate view bias by the reference image and the Janus problem, we incorporate additional multi-view regularization by jointly optimizing $\nabla_\theta \mathcal{L}_{\text{ISD}}$ with a multi-view prior $\nabla_\theta \mathcal{L}_{\text{SDS-MVD}}$. 
An overview of our method is shown in Figure~\ref{fig:pipeline}.

\subsection{Mode guidance using IP-Adapter}
\label{sec:method_IP_sec}
To implement mode guidance from a prior $p_{\phi}(z_t|y,x_{\text{ref}})$ with image prompt capability, we leverage IP-Adapter~\cite{ipadapter}, a lightweight adapter model built upon a pretrained LDM model. To begin, we sample a reference image from an arbitrary text-to-image model, denoted as $x_{\text{ref}}$. Next, we extract a visual token of length $L$ by passing $x_{\text{ref}}$ through an image encoder. This encoder effectively transforms the visual information into a latent that can be utilized in conjunction with text prompt. 
The score distillation using the IP-adapter guidance can be formulated with the IP-SDS loss as:
\begin{align} \label{eq:ip_sds}
    \nabla_\theta \mathcal{L}_{\text{IP-SDS}} \triangleq \mathbb{E}_{t, \epsilon,c} \left[\omega(t)(\epsilon_{\text{IP}} (x_t, t, y, x_{\text{ref}})-\epsilon) \frac{\partial g(\theta, c)}{\partial \theta}\right],
\end{align}
where $\epsilon_{\text{IP}}$ is the noise prediction using IP-adapter. 
As depicted in \cref{fig:sds_3d}, by employing $p_{\phi}(z_t|y,x_{\text{ref}})$ as a teacher, the visual information provided by $x_{\text{ref}}$ guides the optimization to converge to a desirable mode, resulting in the improved 3D quality and training stability compared to the original SDS loss. 

\subsection{Improved control variates with IP-Adapter}

An additional benefit of using IP-Adapter for mode guidance is that we can use it to build a strong control variate that further reduces variances in score estimates.
Let us recall the derivative of the original SDS loss (see appendix A.4 of \cite{poole2023dreamfusion}). For SDS, the gradient can be derived from \cref{eq:sds_loss_form}, which is equal to:
\begin{align} \label{eq:sds_true_grad_expand}
     \mathbb{E}_{\epsilon} \left[ \underbrace{\nabla_{\theta} \log q(z_t | \mathbf{x} = g(\theta))}_{\text{(A)}} - \underbrace{\nabla_{\theta} \log p_{\phi}(z_t | y)}_{\text{(B)}} \right],
\end{align}
where term (B) represents the score function estimated by the LDM' Unet, while term (A) comprises two components -- path derivatives and parameter score -- that sum to zero, yielding the true gradient of SDS:

\begin{align} \label{eq:sds_true_grad}
    \nabla_\theta \mathcal{L}_{\text{SDS}} \triangleq \mathbb{E}_{t, \epsilon,c} \left[\omega(t)(\epsilon_{\theta} (x_t, t, y)) \frac{\partial g(\theta, c)}{\partial \theta}\right],
\end{align}
However, in practice, SDS chooses to keep the path derivative (ie. $\epsilon$) and only discard the parameter score, as the path derivative can be positively correlated to other terms in (B). This leads to \cref{eq:sds}, where $\epsilon$ is the control variate for the prediction function. Here we introduce a reduced variance version of the gradient as follows:
\begin{multline}
    \nabla_\theta \mathcal{L}_{\text{ISD}} \triangleq \mathbb{E}_{t, \epsilon, c} \biggl[\omega(t)
    \Bigl(\epsilon_{\text{IP}}(x_t, t, y, x_\text{ref}) \\ - \epsilon_{\text{SD}}(x_t, t, y)\Bigr) 
    \frac{\partial g(\theta, c)}{\partial \theta} \biggr],
    \label{eq:isd}
\end{multline}
which establishes our \emph{image prompt score distillation} (ISD). 
The rationale behind ISD is that the noise prediction of the Unet without image prompt $\epsilon_{\text{SD}}$ can be closely aligned with those of the IP-Adapter since both still shares the same text prompt, which makes them correlate more positively than with a random noise estimate. 

We found that the variance of the gradients is significantly reduced compared to the pure noise scenario as demonstrated in a 2D toy experiment for 10 prompts using IP-SDS, IP-VSD, and our proposed ISD method. The IP-VSD variant is the application of VSD loss~\cite{wang2023prolificdreamer} on the IP-Adapter prior instead of the original text-to-image prior.
We recorded and calculated the gradient norm for each method, as shown in \cref{fig:gradient grad}.
Overall, the gradient norms of IP-VSD and IP-SDS display similar fluctuation patterns; however, IP-SDS exhibits less oscillation in the gradient norm, indicating that our approach more effectively controls SDS gradient variance when guided by the IP-Adapter. Remarkably, ISD yields results almost identical to IP-VSD in both 2D and 3D generations, as seen in \cref{fig:gradient grad} and \cref{fig:control_variate}. This finding suggests that, with the IP-Adapter as a teacher, the original Unet can effectively work as a control variate, eliminating the need for additional fine-tuning of the LoRA module in IP-VSD.

\subsection{Multi-view regularization}
\label{multiview_regularization}
\begin{figure*}[!htp]
  \centering
  \includegraphics[trim={0, 0.7cm, 0, 0.2cm}, clip,width=1.\linewidth]{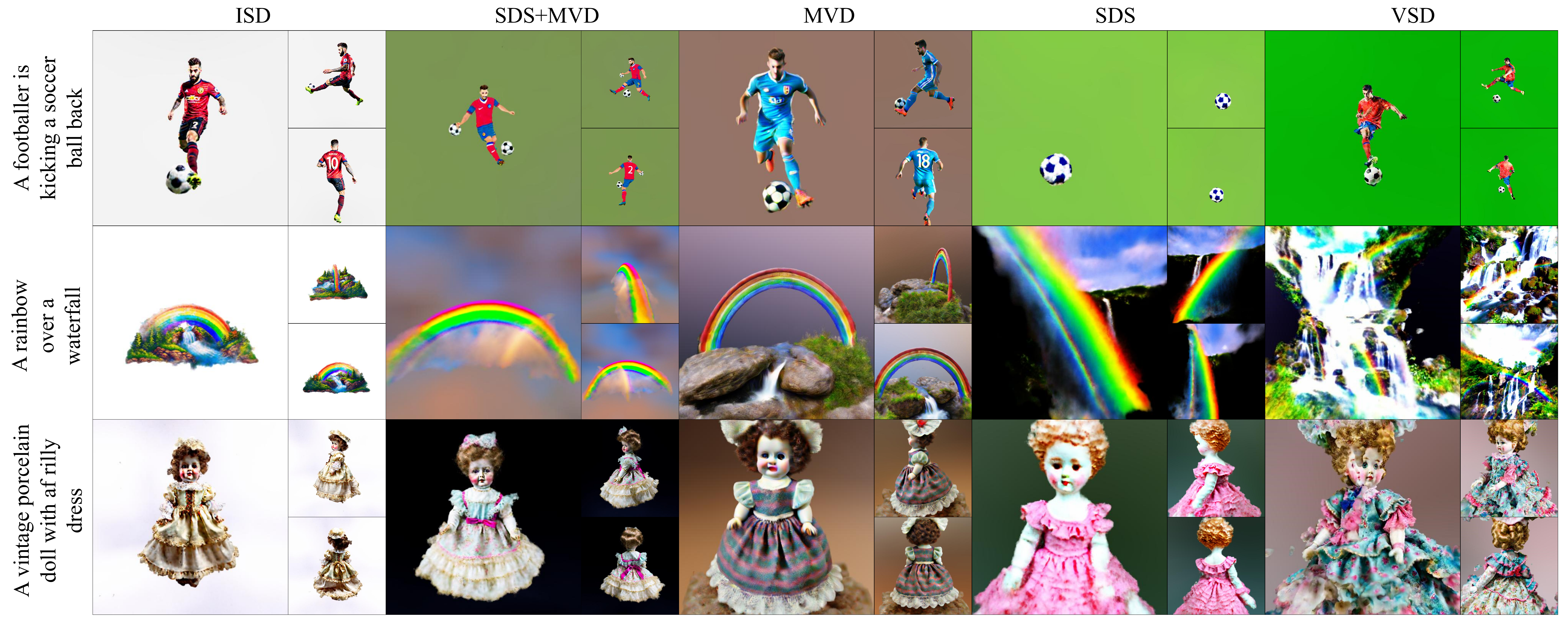}
   \caption{Qualitative comparison between our approach and prior methods including MVDream, SDS, and VSD.}
   \label{fig:qualitative_comparision}
   \vspace{-8pt}
\end{figure*}

A particular issue of using an image prompt to guide score distillation is that the model can be biased toward the reference image views, leading to the Janus problem in the output. To address this, we further leverage MVDream~\cite{shi2024mvdream} as an additional multi-view regularization for score distillation sampling: 
\begin{align} \label{eq:sds-mvd}
    \nabla_\theta \mathcal{L}_{\text{SDS-MVD}} \triangleq \mathbb{E}_{t, \epsilon,c} \left[\omega(t)(\epsilon_{\text{MVD}} (x_t, t, y)-\epsilon) \frac{\partial g(\theta, c)}{\partial \theta}\right],
\end{align}
where $\epsilon_{\text{MVD}}$ is the noise prediction of the MVDream model.
Our combined score distillation is formulated as:
\begin{align} \label{eq:final_loss}
    \nabla_\theta \mathcal{L}_{\text{final}}= \alpha \nabla_\theta \mathcal{L}_{\text{ISD}} + \beta\nabla_\theta \mathcal{L}_{\text{SDS-MVD}},
\end{align}
where $\alpha$ and $\beta$ are two scaling factors. During optimization, $\beta$ gradually decreases to emphasize shape generation in the early iterations, while $\alpha$ gradually increases, shifting the focus toward texture rather than geometry as optimization progresses. The whole process is illustrated in \cref{fig:pipeline}. Both $\alpha$ and $\beta$ are essential components of our methodology. If only $\alpha$ is provided, the resulting 3D outputs will be adversely affected by the Janus problem. Conversely, if only $\beta$ is utilized, the generated 3D object will exhibit a smooth and toy-like texture, as illustrated in \cref{fig:qualitative_comparision}.

\section{Experiments}
\label{sec:experiments}

\begin{table*}[t]
    \small
    \centering
    \setlength{\tabcolsep}{5pt} 
    \begin{tabular}{lcccccccccc}
        \toprule
        & \textbf{Time} & \multicolumn{3}{c}{\textbf{Single Object}} & \multicolumn{3}{c}{\textbf{Single Object with Surr}} & \multicolumn{3}{c}{\textbf{Multiple Objects}} \\ 
        \cmidrule{3-11}
        \textbf{Method} & (mins) & \textbf{Qual.} $\uparrow$ & \textbf{Align.} $\uparrow$ & \textbf{Avg} $\uparrow$
               & \textbf{Qual.} $\uparrow$ & \textbf{Align.} $\uparrow$ & \textbf{Avg} $\uparrow$
               & \textbf{Qual.} $\uparrow$ & \textbf{Align.} $\uparrow$ & \textbf{Avg} $\uparrow$\\ 
        \midrule
        Dreamfusion \cite{poole2023dreamfusion} & 30 & 24.9 & 24.0 & 24.4 & 19.3 & 29.8 & 24.6 & 17.3 & 14.8 & 16.1 \\ 
        Magic3D \cite{magic3d} & 40 & 38.7 & 35.3 & 37.0 & 29.8 & 41.0 & 35.4 & 26.6 & 24.8 & 25.7 \\ 
        LatentNeRF \cite{Metzer_2023_CVPR} & 65 & 34.2 & 32.0 & 33.1 & 23.7 & 37.5 & 30.6 & 21.7 & 19.5 & 20.6 \\ 
        Fantasia3D \cite{fantasia3d} & 45 & 29.2 & 23.5 & 26.4 & 21.9 & 32.0 & 27.0 & 22.7 & 14.3 & 18.5 \\ 
        SJC \cite{sjc} & 25 & 26.3 & 23.0 & 24.7 & 17.3 & 22.3 & 19.8 & 11.7 & 5.8 & 8.7 \\ 
        ProlificDreamer \cite{wang2023prolificdreamer} & 240 & 51.1 & \underline{47.8} & \underline{49.4} & 42.5 & 47.0 & \underline{44.8} & \textbf{45.7} & 25.8 & \underline{35.8} \\ 
        MVDream \cite{shi2024mvdream} & 30 & 53.2 & 42.3 & 47.8 & 36.3 & \underline{48.5} & 42.4 & 39.0 & \underline{28.5} & 33.8 \\ 
        DreamGaussian \cite{tang2024dreamgaussian} & 7 & 19.9 & 19.8 & 19.8 & 10.4 & 17.8 & 14.1 & 12.3 & 9.5 & 10.9 \\ 
        GeoDream \cite{Ma2023GeoDream} & 400 & 48.4 & 33.8 & 41.1 & 35.2 & 34.5 & 34.9 & 34.3 & 16.5 & 25.4 \\ 
        RichDreamer \cite{qiu2024richdreamer} & 70 & \textbf{57.3} & 40.0 & 48.6 & \underline{43.9} & 42.3 & 43.1 & 34.8 & 22.0 & 28.4 \\ 
        \midrule
        ISD (ours) & 40 & \underline{55.4} & \textbf{52.6} & \textbf{54.0} & \textbf{45.7} & \textbf{59.0} & \textbf{52.4} & \underline{43.4} & \textbf{39.4} & \textbf{41.4} \\
        \bottomrule
    \end{tabular}
    \caption{Comparative results for the text-to-3D task across three settings of T3Bench. The best results are \textbf{bold} while the second best results are \underline{underlined}.}
    \label{tab:t3bench_results}
    \vspace{-5pt}
\end{table*}

\begin{figure*}[!htp]
  \centering
  \includegraphics[width=1.\linewidth]{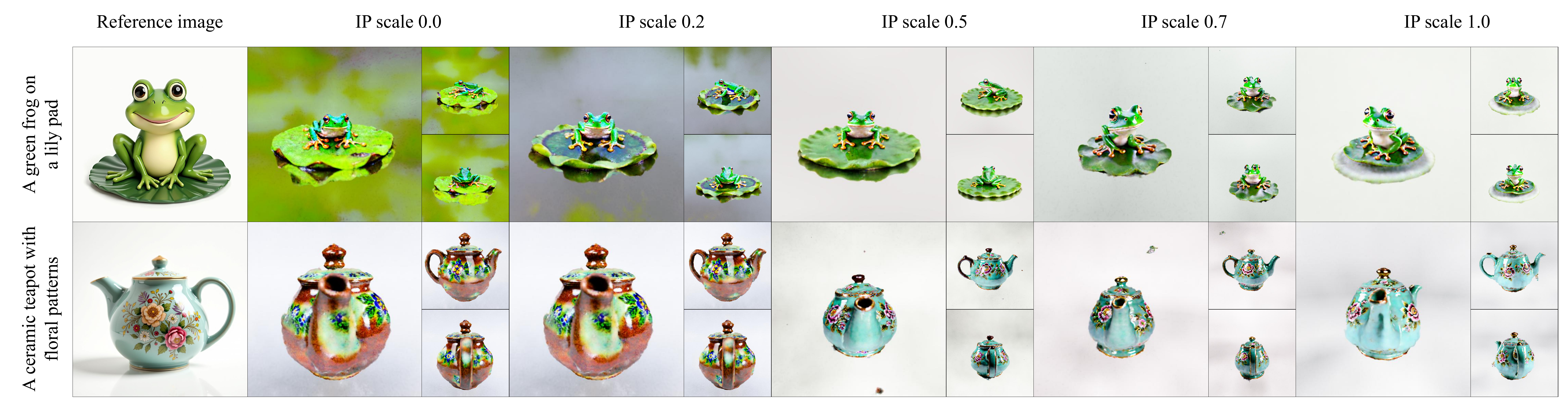}
   \caption{The effect of the IP scale indicates that as the IP scale increases, the generated 3D results align more closely with the reference image; however, this comes at the cost of worsening the Janus problem.}
   \label{fig:qualitative_ablation_ip_scale}
  \vspace{-10pt}
\end{figure*}

\myheading{Benchmark and metrics}: We evaluate our method using T3bench \cite{t3bench}, a benchmark that assesses both the quality and text prompt alignment of generated 3D models. T3bench consists of 300 prompts, categorized into three groups: single object, single object with surroundings, and multiple objects.
For qualitative evaluation, T3bench converts a 3D representation into a mesh and defines a level-0 icosahedron, with vertices representing camera positions for multi-view image rendering. These images are processed by ImageReward \cite{image_reward}, and the results are aggregated to generate a quality score.
For text prompt alignment, each multi-view image is captioned using BLIP \cite{li2022blip}. These captions, along with those of 3D models generated by CAP3D \cite{cap3D}, are merged by ChatGPT and compared to the initial prompt.

\myheading{Implementation details}: Our method is implemented based on the ThreeStudio framework \cite{threestudio2023}. We train each 3D object for 6,000 iterations, using the IP-Adapter \cite{ipadapter} version SD 1.5 as an approximation for $p_{\phi}(z_t|y,x_{\text{ref}})$, with the IP scale set to 0.5. We also employ MVDream \cite{shi2024mvdream} as a multi-view regularization technique to address the Janus problem. The scaling factors $\alpha$ and $\beta$ in \cref{eq:final_loss} are initially set to 0.4 and 0.6, respectively, with $\alpha$ increasing to 0.8 and $\beta$ decreasing to 0.02 as training progresses. We employed a classifier-free guidance scale (CFG) of 7.5 for the ISD term and a CFG of 50 for the SDS-MVD regularization.

\subsection{Comparison with prior methods}
\myheading{Quantitative Comparison}:
We use the methods presented in T3bench \cite{t3bench} as our comparison baseline, including DreamFusion \cite{poole2023dreamfusion}, Magic3D \cite{magic3d}, LatentNeRF \cite{Metzer_2023_CVPR}, Fantasia3D \cite{fantasia3d}, SJC \cite{sjc}, ProlificDreamer \cite{wang2023prolificdreamer}, MVDream \cite{shi2024mvdream}, DreamGaussian \cite{tang2024dreamgaussian}, GeoDream \cite{Ma2023GeoDream}, and RichDreamer \cite{qiu2024richdreamer}. Quantitative results are presented in \cref{tab:t3bench_results}. In the single-object category, our method achieves near SOTA performance, comparable to RichDreamer \cite{qiu2024richdreamer}, while cutting training time nearly in half, surpassing previous methods. For single objects with surroundings, we attain the highest benchmark score. In the multiple-objects category, our method ranks second only to VSD, with a small margin, due to MVDream’s geometric limitations. However, by leveraging guidance from reference images, our method improves 3D model quality over raw MVDream \cite{shi2024mvdream}. Notably, our method achieves the highest alignment score across all categories due to effective reference image guidance.

\myheading{Qualitative Comparison.} We present comparisons with prior methods in \cref{fig:qualitative_comparision}. While SDS struggles with poor texture and geometry, VSD improves texture quality but still shows weak geometry. MVDream achieves reasonable geometry but suffers from toy-like textures due to its bias toward the Objaverse \cite{objaverse} dataset. In contrast, our method combines the strengths of both approaches, generating realistic geometry with MVDream’s constraints while achieving high-quality texture by leveraging the mode selected in the reference image.

\subsection{Ablation study}

\begin{table}[t]
\centering
\footnotesize
\setlength{\tabcolsep}{1pt}
\begin{minipage}{0.22\textwidth}
    \centering
    \begin{tabular}{cccl}
        \toprule
        $\textbf{IP Scale}$ & \textbf{Qual.} $\uparrow$ & \textbf{Align.} $\uparrow$ \\
        \midrule
        0.0  & 69.2 & 53.96 \\
        0.2  & 69.4 & 54.76 \\
        0.5  & \textbf{81.9} & 54.89\\
        0.7  & 76.8 & \textbf{55.21} \\
        1.0  & 70.9 & 55.18\\
        \bottomrule
    \end{tabular}
    \caption{Ablation study of different IP scales.}
    \label{tab:quantitative_ip_scale}
    \vspace{-10pt}
\end{minipage}%
\hfill
\begin{minipage}{0.25\textwidth}
    \centering
    \begin{tabular}{cccl}
        \toprule
        $\textbf{Control variate}$ & \textbf{Qual.} $\uparrow$ & \textbf{Align.} $\uparrow$ \\
        \midrule
        Rand. noise (CFG 7.5)  & 45.8 & 54.65\\
        Rand. noise (CFG 100)  & 75.7 & 54.82 \\
        Unet LoRA  & 79.6 & \textbf{54.96}\\
        Unet ($t + \delta t$)  & 72.9 & 54.52 \\
        \midrule
        Unet (same $t$) (ours)  & \textbf{81.8} & 54.89 \\
        \bottomrule
    \end{tabular}
    \caption{Ablation study of different control variate elements.}
    \label{tab:quantitative_control_var}
    \vspace{-10pt}
\end{minipage}
\end{table}

In this section, we conduct several experiments to identify potential factors affecting our method, including the variance of the IP scale and various variance control strategies. We perform an ablation study using 20 prompts selected from three categories of the T3bench benchmark.

\myheading{Effect of IP scale.} Using IP-Adapter, the IP scale adjusts the influence of image prompting. We vary the IP scale across values [0.0, 0.2, 0.5, 0.7, 1.0] to evaluate its effect on our methods. Quantitative results are provided in \cref{tab:quantitative_ip_scale}, and qualitative results in \cref{fig:qualitative_ablation_ip_scale}. As shown, higher IP scales produce 3D objects that align more closely with the reference images but are also more prone to the Janus problem. We select an IP scale of 0.5 to balance reference image alignment with a reduced likelihood of the Janus problem.

\myheading{Ablation of variate control.} Another key factor in our method is the choice of control variate, represented by the second term in \cref{eq:isd}. In this study, we test several alternatives: replacing the $\text{Unet}$ term with pure noise (IP-SDS), substituting Unet with a trainable LoRA (IP-VSD), and using Unet at $t + \delta t$ instead of time $t$. Both qualitative results in \cref{fig:control_variate} and quantitative results in \cref{tab:quantitative_control_var} show that using $\text{Unet}$ as the control variate achieves results similar to Unet with LoRA, allowing our method to approximate IP-VSD without the additional LoRA training, thereby significantly speeding up training. Additionally, the Gaussian noise control variate requires a larger CFG value of $100$ to yield coherent 3D objects, which notably raises the likelihood of the Janus problem. When using Unet at $t + \delta t$, the 3D quality is generally similar to that with our Unet at $t$, but some generated objects display missing components. Thus, we select Unet at $t$ as our default control variate, achieving an effective balance of generation speed and 3D quality.

\section{Limitations and Conclusion}
\label{sec:conclusion}

\myheading{Limitations}: Although our method can generate high-quality 3D objects, it still suffers from the Janus problem due to biases in the views of the input images provided by the IP-Adapter. Additionally, our method is constrained by the performance of both the IP-Adapter and MVDream, which hinders its ability to generate composite objects. Another drawback is that the generated geometry may not always align with the reference image, resulting in significant geometric shifts in later iterations.

\myheading{Conclusion}: In this paper, we have introduced a novel framework for selecting a high-quality mode prior for text-to-3D generation. By leveraging the mode suggested by a reference image and a better control variate to reduce variance in score estimate, our method can generate high-quality 3D models at a reasonable speed. In future work, we aim to explore more effective mitigation strategies for the Janus problem when utilizing the IP-Adapter as a mode selector. Additionally, we plan to extend our methods to various settings such as amortized optimization and for other 3D representations.

{
    \small
    \bibliographystyle{ieeenat_fullname}
    \bibliography{main}

\begin{thebibliography}{66}
\providecommand{\natexlab}[1]{#1}
\providecommand{\url}[1]{\texttt{#1}}
\expandafter\ifx\csname urlstyle\endcsname\relax
  \providecommand{\doi}[1]{doi: #1}\else
  \providecommand{\doi}{doi: \begingroup \urlstyle{rm}\Url}\fi

\bibitem[Anonymous(2024)]{anonymous2024eliminating}
Anonymous.
\newblock Eliminating oversaturation and artifacts of high guidance scales in diffusion models.
\newblock In \emph{Submitted to The Thirteenth International Conference on Learning Representations}, 2024.
\newblock under review.

\bibitem[Chen et~al.(2023)Chen, Chen, Jiao, and Jia]{fantasia3d}
Rui Chen, Yongwei Chen, Ningxin Jiao, and Kui Jia.
\newblock Fantasia3d: Disentangling geometry and appearance for high-quality text-to-3d content creation.
\newblock In \emph{Proceedings of the IEEE/CVF International Conference on Computer Vision (ICCV)}, 2023.

\bibitem[Chen et~al.(2024)Chen, Pan, Yang, Yao, and Mei]{Yang2024VP3D}
Yang Chen, Yingwei Pan, Haibo Yang, Ting Yao, and Tao Mei.
\newblock Vp3d: Unleashing 2d visual prompt for text-to-3d generation.
\newblock In \emph{CVPR}, 2024.

\bibitem[Deitke et~al.(2023{\natexlab{a}})Deitke, Liu, Wallingford, Ngo, Michel, Kusupati, Fan, Laforte, Voleti, Gadre, VanderBilt, Kembhavi, Vondrick, Gkioxari, Ehsani, Schmidt, and Farhadi]{objaverseXL}
Matt Deitke, Ruoshi Liu, Matthew Wallingford, Huong Ngo, Oscar Michel, Aditya Kusupati, Alan Fan, Christian Laforte, Vikram Voleti, Samir~Yitzhak Gadre, Eli VanderBilt, Aniruddha Kembhavi, Carl Vondrick, Georgia Gkioxari, Kiana Ehsani, Ludwig Schmidt, and Ali Farhadi.
\newblock Objaverse-xl: A universe of 10m+ 3d objects.
\newblock \emph{arXiv preprint arXiv:2307.05663}, 2023{\natexlab{a}}.

\bibitem[Deitke et~al.(2023{\natexlab{b}})Deitke, Schwenk, Salvador, Weihs, Michel, VanderBilt, Schmidt, Ehsani, Kembhavi, and Farhadi]{objaverse}
Matt Deitke, Dustin Schwenk, Jordi Salvador, Luca Weihs, Oscar Michel, Eli VanderBilt, Ludwig Schmidt, Kiana Ehsani, Aniruddha Kembhavi, and Ali Farhadi.
\newblock Objaverse: A universe of annotated 3d objects.
\newblock In \emph{Proceedings of the IEEE/CVF Conference on Computer Vision and Pattern Recognition}, pages 13142--13153, 2023{\natexlab{b}}.

\bibitem[Ding et~al.(2024)Ding, Dong, Huang, Wang, Zhang, Gong, Xu, and Xue]{ding2024text}
Lihe Ding, Shaocong Dong, Zhanpeng Huang, Zibin Wang, Yiyuan Zhang, Kaixiong Gong, Dan Xu, and Tianfan Xue.
\newblock Text-to-3d generation with bidirectional diffusion using both 2d and 3d priors.
\newblock In \emph{Proceedings of the IEEE/CVF Conference on Computer Vision and Pattern Recognition}, pages 5115--5124, 2024.

\bibitem[Esser et~al.(2024)Esser, Kulal, Blattmann, Entezari, M{\"u}ller, Saini, Levi, Lorenz, Sauer, Boesel, et~al.]{esser2024scaling}
Patrick Esser, Sumith Kulal, Andreas Blattmann, Rahim Entezari, Jonas M{\"u}ller, Harry Saini, Yam Levi, Dominik Lorenz, Axel Sauer, Frederic Boesel, et~al.
\newblock Scaling rectified flow transformers for high-resolution image synthesis.
\newblock In \emph{Forty-first International Conference on Machine Learning}, 2024.

\bibitem[Goodfellow et~al.(2014)Goodfellow, Pouget-Abadie, Mirza, Xu, Warde-Farley, Ozair, Courville, and Bengio]{goodfellow2014generative}
Ian Goodfellow, Jean Pouget-Abadie, Mehdi Mirza, Bing Xu, David Warde-Farley, Sherjil Ozair, Aaron Courville, and Yoshua Bengio.
\newblock Generative adversarial nets.
\newblock \emph{Advances in neural information processing systems}, 27, 2014.

\bibitem[Guo et~al.(2023)Guo, Liu, Shao, Laforte, Voleti, Luo, Chen, Zou, Wang, Cao, and Zhang]{threestudio2023}
Yuan-Chen Guo, Ying-Tian Liu, Ruizhi Shao, Christian Laforte, Vikram Voleti, Guan Luo, Chia-Hao Chen, Zi-Xin Zou, Chen Wang, Yan-Pei Cao, and Song-Hai Zhang.
\newblock threestudio: A unified framework for 3d content generation, 2023.

\bibitem[He et~al.(2023{\natexlab{a}})He, Bai, Lin, Zhao, Hu, Sheng, Yi, Li, and Liu]{he2023t3bench}
Yuze He, Yushi Bai, Matthieu Lin, Wang Zhao, Yubin Hu, Jenny Sheng, Ran Yi, Juanzi Li, and Yong-Jin Liu.
\newblock T$^3$bench: Benchmarking current progress in text-to-3d generation, 2023{\natexlab{a}}.

\bibitem[He et~al.(2023{\natexlab{b}})He, Bai, Lin, Zhao, Hu, Sheng, Yi, Li, and Liu]{t3bench}
Yuze He, Yushi Bai, Matthieu Lin, Wang Zhao, Yubin Hu, Jenny Sheng, Ran Yi, Juanzi Li, and Yong-Jin Liu.
\newblock T$^3$bench: Benchmarking current progress in text-to-3d generation, 2023{\natexlab{b}}.

\bibitem[Hong et~al.(2024)Hong, Zhang, Gu, Bi, Zhou, Liu, Liu, Sunkavalli, Bui, and Tan]{hong2024lrm}
Yicong Hong, Kai Zhang, Jiuxiang Gu, Sai Bi, Yang Zhou, Difan Liu, Feng Liu, Kalyan Sunkavalli, Trung Bui, and Hao Tan.
\newblock {LRM}: Large reconstruction model for single image to 3d.
\newblock In \emph{The Twelfth International Conference on Learning Representations}, 2024.

\bibitem[Hu et~al.(2022)Hu, Shen, Wallis, Allen-Zhu, Li, Wang, Wang, and Chen]{hu2021lora}
Edward~J Hu, Yelong Shen, Phillip Wallis, Zeyuan Allen-Zhu, Yuanzhi Li, Shean Wang, Lu Wang, and Weizhu Chen.
\newblock Lo{RA}: Low-rank adaptation of large language models.
\newblock In \emph{International Conference on Learning Representations}, 2022.

\bibitem[Huang et~al.(2024)Huang, Wang, Shi, Tang, Qi, and Zhang]{huang2024dreamtime}
Yukun Huang, Jianan Wang, Yukai Shi, Boshi Tang, Xianbiao Qi, and Lei Zhang.
\newblock Dreamtime: An improved optimization strategy for diffusion-guided 3d generation.
\newblock In \emph{The Twelfth International Conference on Learning Representations}, 2024.

\bibitem[Jain et~al.(2022)Jain, Mildenhall, Barron, Abbeel, and Poole]{dreamfields}
Ajay Jain, Ben Mildenhall, Jonathan~T Barron, Pieter Abbeel, and Ben Poole.
\newblock Zero-shot text-guided object generation with dream fields.
\newblock In \emph{Proceedings of the IEEE/CVF Conference on Computer Vision and Pattern Recognition}, pages 867--876, 2022.

\bibitem[Jiang et~al.(2025)Jiang, Zeng, Hu, Xu, Zhang, Xu, and Yeung]{jiang2025jointdreamer}
Chenhan Jiang, Yihan Zeng, Tianyang Hu, Songcun Xu, Wei Zhang, Hang Xu, and Dit-Yan Yeung.
\newblock Jointdreamer: Ensuring geometry consistency and text congruence in text-to-3d generation via joint score distillation.
\newblock In \emph{European Conference on Computer Vision}, pages 439--456. Springer, 2025.

\bibitem[Jun and Nichol(2023)]{shap-e}
Heewoo Jun and Alex Nichol.
\newblock Shap-e: Generating conditional 3d implicit functions, 2023.

\bibitem[Katzir et~al.(2024)Katzir, Patashnik, Cohen-Or, and Lischinski]{katzir2024noisefree}
Oren Katzir, Or Patashnik, Daniel Cohen-Or, and Dani Lischinski.
\newblock Noise-free score distillation.
\newblock In \emph{The Twelfth International Conference on Learning Representations}, 2024.

\bibitem[Kerbl et~al.(2023)Kerbl, Kopanas, Leimk{\"u}hler, and Drettakis]{kerbl3Dgaussians}
Bernhard Kerbl, Georgios Kopanas, Thomas Leimk{\"u}hler, and George Drettakis.
\newblock 3d gaussian splatting for real-time radiance field rendering.
\newblock \emph{ACM Transactions on Graphics}, 42\penalty0 (4), 2023.

\bibitem[Lee et~al.(2024)Lee, Sohn, and Shin]{lee2024dreamflow}
Kyungmin Lee, Kihyuk Sohn, and Jinwoo Shin.
\newblock Dreamflow: High-quality text-to-3d generation by approximating probability flow.
\newblock In \emph{The Twelfth International Conference on Learning Representations}, 2024.

\bibitem[Li et~al.(2022)Li, Li, Xiong, and Hoi]{li2022blip}
Junnan Li, Dongxu Li, Caiming Xiong, and Steven Hoi.
\newblock Blip: Bootstrapping language-image pre-training for unified vision-language understanding and generation.
\newblock In \emph{ICML}, 2022.

\bibitem[Li et~al.(2024{\natexlab{a}})Li, Tan, Zhang, Xu, Luan, Xu, Hong, Sunkavalli, Shakhnarovich, and Bi]{li2024instantd}
Jiahao Li, Hao Tan, Kai Zhang, Zexiang Xu, Fujun Luan, Yinghao Xu, Yicong Hong, Kalyan Sunkavalli, Greg Shakhnarovich, and Sai Bi.
\newblock Instant3d: Fast text-to-3d with sparse-view generation and large reconstruction model.
\newblock In \emph{The Twelfth International Conference on Learning Representations}, 2024{\natexlab{a}}.

\bibitem[Li et~al.(2024{\natexlab{b}})Li, Liu, Chen, Liang, Chen, Tan, and Long]{li2024craftsman}
Weiyu Li, Jiarui Liu, Rui Chen, Yixun Liang, Xuelin Chen, Ping Tan, and Xiaoxiao Long.
\newblock Craftsman: High-fidelity mesh generation with 3d native generation and interactive geometry refiner.
\newblock \emph{arXiv preprint arXiv:2405.14979}, 2024{\natexlab{b}}.

\bibitem[Lin et~al.(2023)Lin, Gao, Tang, Takikawa, Zeng, Huang, Kreis, Fidler, Liu, and Lin]{magic3d}
Chen-Hsuan Lin, Jun Gao, Luming Tang, Towaki Takikawa, Xiaohui Zeng, Xun Huang, Karsten Kreis, Sanja Fidler, Ming-Yu Liu, and Tsung-Yi Lin.
\newblock Magic3d: High-resolution text-to-3d content creation.
\newblock In \emph{IEEE Conference on Computer Vision and Pattern Recognition ({CVPR})}, 2023.

\bibitem[Liu et~al.(2023{\natexlab{a}})Liu, Xu, Jin, Chen, T, Xu, and Su]{one2345}
Minghua Liu, Chao Xu, Haian Jin, Linghao Chen, Mukund~Varma T, Zexiang Xu, and Hao Su.
\newblock One-2-3-45: Any single image to 3d mesh in 45 seconds without per-shape optimization.
\newblock In \emph{Thirty-seventh Conference on Neural Information Processing Systems}, 2023{\natexlab{a}}.

\bibitem[Liu et~al.(2024{\natexlab{a}})Liu, Xu, Jin, Chen, Varma~T, Xu, and Su]{liu2023one2345}
Minghua Liu, Chao Xu, Haian Jin, Linghao Chen, Mukund Varma~T, Zexiang Xu, and Hao Su.
\newblock One-2-3-45: Any single image to 3d mesh in 45 seconds without per-shape optimization.
\newblock \emph{Advances in Neural Information Processing Systems}, 36, 2024{\natexlab{a}}.

\bibitem[Liu and Wang(2016)]{liu2016stein}
Qiang Liu and Dilin Wang.
\newblock Stein variational gradient descent: A general purpose bayesian inference algorithm.
\newblock \emph{Advances in neural information processing systems}, 29, 2016.

\bibitem[Liu et~al.(2023{\natexlab{b}})Liu, Wu, Van~Hoorick, Tokmakov, Zakharov, and Vondrick]{zero123}
Ruoshi Liu, Rundi Wu, Basile Van~Hoorick, Pavel Tokmakov, Sergey Zakharov, and Carl Vondrick.
\newblock Zero-1-to-3: Zero-shot one image to 3d object.
\newblock In \emph{Proceedings of the IEEE/CVF International Conference on Computer Vision (ICCV)}, pages 9298--9309, 2023{\natexlab{b}}.

\bibitem[Liu et~al.(2024{\natexlab{b}})Liu, Lin, Zeng, Long, Liu, Komura, and Wang]{liu2024syncdreamer}
Yuan Liu, Cheng Lin, Zijiao Zeng, Xiaoxiao Long, Lingjie Liu, Taku Komura, and Wenping Wang.
\newblock Syncdreamer: Generating multiview-consistent images from a single-view image.
\newblock In \emph{The Twelfth International Conference on Learning Representations}, 2024{\natexlab{b}}.

\bibitem[Lorraine et~al.(2023)Lorraine, Xie, Zeng, Lin, Takikawa, Sharp, Lin, Liu, Fidler, and Lucas]{lorraine2023att3d}
Jonathan Lorraine, Kevin Xie, Xiaohui Zeng, Chen-Hsuan Lin, Towaki Takikawa, Nicholas Sharp, Tsung-Yi Lin, Ming-Yu Liu, Sanja Fidler, and James Lucas.
\newblock Att3d: Amortized text-to-3d object synthesis.
\newblock \emph{The International Conference on Computer Vision (ICCV)}, 2023.

\bibitem[Luo et~al.(2023)Luo, Rockwell, Lee, and Johnson]{cap3D}
Tiange Luo, Chris Rockwell, Honglak Lee, and Justin Johnson.
\newblock Scalable 3d captioning with pretrained models.
\newblock In \emph{Advances in Neural Information Processing Systems}, pages 75307--75337. Curran Associates, Inc., 2023.

\bibitem[Ma et~al.(2023)Ma, Deng, Zhou, Liu, Huang, and Wang]{Ma2023GeoDream}
Baorui Ma, Haoge Deng, Junsheng Zhou, Yu-Shen Liu, Tiejun Huang, and Xinlong Wang.
\newblock Geodream: Disentangling 2d and geometric priors for high-fidelity and consistent 3d generation.
\newblock \emph{arXiv preprint arXiv:2311.17971}, 2023.

\bibitem[Ma et~al.(2024)Ma, Wei, Zhang, Zhu, Lei, and Zhang]{ma2024scaledreamer}
Zhiyuan Ma, Yuxiang Wei, Yabin Zhang, Xiangyu Zhu, Zhen Lei, and Lei Zhang.
\newblock Scaledreamer: Scalable text-to-3d synthesis with asynchronous score distillation.
\newblock \emph{The 18th European Conference on Computer Vision (ECCV)}, 2024.

\bibitem[Melas-Kyriazi et~al.(2023)Melas-Kyriazi, Laina, Rupprecht, and Vedaldi]{realfusion}
Luke Melas-Kyriazi, Iro Laina, Christian Rupprecht, and Andrea Vedaldi.
\newblock Realfusion: 360deg reconstruction of any object from a single image.
\newblock In \emph{Proceedings of the IEEE/CVF Conference on Computer Vision and Pattern Recognition (CVPR)}, pages 8446--8455, 2023.

\bibitem[Metzer et~al.(2023)Metzer, Richardson, Patashnik, Giryes, and Cohen-Or]{Metzer_2023_CVPR}
Gal Metzer, Elad Richardson, Or Patashnik, Raja Giryes, and Daniel Cohen-Or.
\newblock Latent-nerf for shape-guided generation of 3d shapes and textures.
\newblock In \emph{Proceedings of the IEEE/CVF Conference on Computer Vision and Pattern Recognition (CVPR)}, pages 12663--12673, 2023.

\bibitem[Mildenhall et~al.(2020)Mildenhall, Srinivasan, Tancik, Barron, Ramamoorthi, and Ng]{mildenhall2020nerf}
Ben Mildenhall, Pratul~P. Srinivasan, Matthew Tancik, Jonathan~T. Barron, Ravi Ramamoorthi, and Ren Ng.
\newblock Nerf: Representing scenes as neural radiance fields for view synthesis.
\newblock In \emph{ECCV}, 2020.

\bibitem[Nichol et~al.(2022)Nichol, Jun, Dhariwal, Mishkin, and Chen]{point-e}
Alex Nichol, Heewoo Jun, Prafulla Dhariwal, Pamela Mishkin, and Mark Chen.
\newblock Point-e: A system for generating 3d point clouds from complex prompts, 2022.

\bibitem[Poole et~al.(2023)Poole, Jain, Barron, and Mildenhall]{poole2023dreamfusion}
Ben Poole, Ajay Jain, Jonathan~T. Barron, and Ben Mildenhall.
\newblock Dreamfusion: Text-to-3d using 2d diffusion.
\newblock In \emph{The Eleventh International Conference on Learning Representations}, 2023.

\bibitem[Qiu et~al.(2024)Qiu, Chen, Gu, Zuo, Xu, Wu, Yuan, Dong, Bo, and Han]{qiu2024richdreamer}
Lingteng Qiu, Guanying Chen, Xiaodong Gu, Qi Zuo, Mutian Xu, Yushuang Wu, Weihao Yuan, Zilong Dong, Liefeng Bo, and Xiaoguang Han.
\newblock Richdreamer: A generalizable normal-depth diffusion model for detail richness in text-to-3d.
\newblock In \emph{Proceedings of the IEEE/CVF Conference on Computer Vision and Pattern Recognition}, pages 9914--9925, 2024.

\bibitem[Ramesh et~al.(2021)Ramesh, Pavlov, Goh, Gray, Voss, Radford, Chen, and Sutskever]{dall-e}
Aditya Ramesh, Mikhail Pavlov, Gabriel Goh, Scott Gray, Chelsea Voss, Alec Radford, Mark Chen, and Ilya Sutskever.
\newblock Zero-shot text-to-image generation.
\newblock In \emph{Proceedings of the 38th International Conference on Machine Learning}, pages 8821--8831. PMLR, 2021.

\bibitem[Rombach et~al.(2022)Rombach, Blattmann, Lorenz, Esser, and Ommer]{stable-diffusion}
Robin Rombach, Andreas Blattmann, Dominik Lorenz, Patrick Esser, and Bj\"orn Ommer.
\newblock High-resolution image synthesis with latent diffusion models.
\newblock In \emph{CVPR}, 2022.

\bibitem[Saharia et~al.(2022)Saharia, Chan, Saxena, Li, Whang, Denton, Ghasemipour, Gontijo-Lopes, Ayan, Salimans, Ho, Fleet, and Norouzi]{saharia2022photorealistic}
Chitwan Saharia, William Chan, Saurabh Saxena, Lala Li, Jay Whang, Emily Denton, Seyed Kamyar~Seyed Ghasemipour, Raphael Gontijo-Lopes, Burcu~Karagol Ayan, Tim Salimans, Jonathan Ho, David~J. Fleet, and Mohammad Norouzi.
\newblock Photorealistic text-to-image diffusion models with deep language understanding.
\newblock In \emph{Advances in Neural Information Processing Systems}, 2022.

\bibitem[Schuhmann et~al.(2022)Schuhmann, Beaumont, Vencu, Gordon, Wightman, Cherti, Coombes, Katta, Mullis, Wortsman, Schramowski, Kundurthy, Crowson, Schmidt, Kaczmarczyk, and Jitsev]{schuhmann2022laion5bopenlargescaledataset}
Christoph Schuhmann, Romain Beaumont, Richard Vencu, Cade Gordon, Ross Wightman, Mehdi Cherti, Theo Coombes, Aarush Katta, Clayton Mullis, Mitchell Wortsman, Patrick Schramowski, Srivatsa Kundurthy, Katherine Crowson, Ludwig Schmidt, Robert Kaczmarczyk, and Jenia Jitsev.
\newblock Laion-5b: An open large-scale dataset for training next generation image-text models, 2022.

\bibitem[Shen et~al.(2024)Shen, Wu, Yi, Zhou, Zhang, Yan, and Wang]{shen2024gamba}
Qiuhong Shen, Zike Wu, Xuanyu Yi, Pan Zhou, Hanwang Zhang, Shuicheng Yan, and Xinchao Wang.
\newblock Gamba: Marry gaussian splatting with mamba for single view 3d reconstruction.
\newblock \emph{arXiv preprint arXiv:2403.18795}, 2024.

\bibitem[Shi et~al.(2023)Shi, Chen, Zhang, Liu, Xu, Wei, Chen, Zeng, and Su]{shi2023zero123plus}
Ruoxi Shi, Hansheng Chen, Zhuoyang Zhang, Minghua Liu, Chao Xu, Xinyue Wei, Linghao Chen, Chong Zeng, and Hao Su.
\newblock Zero123++: a single image to consistent multi-view diffusion base model, 2023.

\bibitem[Shi et~al.(2024)Shi, Wang, Ye, Mai, Li, and Yang]{shi2024mvdream}
Yichun Shi, Peng Wang, Jianglong Ye, Long Mai, Kejie Li, and Xiao Yang.
\newblock {MVD}ream: Multi-view diffusion for 3d generation.
\newblock In \emph{The Twelfth International Conference on Learning Representations}, 2024.

\bibitem[Tang et~al.(2024{\natexlab{a}})Tang, Chen, Chen, Wang, Zeng, and Liu]{tang2024lgm}
Jiaxiang Tang, Zhaoxi Chen, Xiaokang Chen, Tengfei Wang, Gang Zeng, and Ziwei Liu.
\newblock Lgm: Large multi-view gaussian model for high-resolution 3d content creation, 2024{\natexlab{a}}.

\bibitem[Tang et~al.(2024{\natexlab{b}})Tang, Ren, Zhou, Liu, and Zeng]{tang2024dreamgaussian}
Jiaxiang Tang, Jiawei Ren, Hang Zhou, Ziwei Liu, and Gang Zeng.
\newblock Dreamgaussian: Generative gaussian splatting for efficient 3d content creation.
\newblock In \emph{The Twelfth International Conference on Learning Representations}, 2024{\natexlab{b}}.

\bibitem[Tochilkin et~al.(2024)Tochilkin, Pankratz, Liu, Huang, , Letts, Li, Liang, Laforte, Jampani, and Cao]{TripoSR2024}
Dmitry Tochilkin, David Pankratz, Zexiang Liu, Zixuan Huang, , Adam Letts, Yangguang Li, Ding Liang, Christian Laforte, Varun Jampani, and Yan-Pei Cao.
\newblock Triposr: Fast 3d object reconstruction from a single image.
\newblock \emph{arXiv preprint arXiv:2403.02151}, 2024.

\bibitem[Tran et~al.(2025)Tran, Luu, Nguyen, Nguyen, and Hua]{tran2025diverse}
Uy~Dieu Tran, Minh Luu, Phong~Ha Nguyen, Khoi Nguyen, and Binh-Son Hua.
\newblock Diverse text-to-3d synthesis with augmented text embedding.
\newblock In \emph{European Conference on Computer Vision}, pages 217--235. Springer, 2025.

\bibitem[Wang et~al.(2023{\natexlab{a}})Wang, Du, Li, Yeh, and Shakhnarovich]{sjc}
Haochen Wang, Xiaodan Du, Jiahao Li, Raymond~A. Yeh, and Greg Shakhnarovich.
\newblock Score jacobian chaining: Lifting pretrained 2d diffusion models for 3d generation.
\newblock In \emph{Proceedings of the IEEE/CVF Conference on Computer Vision and Pattern Recognition (CVPR)}, pages 12619--12629, 2023{\natexlab{a}}.

\bibitem[Wang and Shi(2023)]{wang2023imagedream}
Peng Wang and Yichun Shi.
\newblock Imagedream: Image-prompt multi-view diffusion for 3d generation.
\newblock \emph{arXiv preprint arXiv:2312.02201}, 2023.

\bibitem[Wang et~al.(2023{\natexlab{b}})Wang, Lu, Wang, Bao, Li, Su, and Zhu]{wang2023prolificdreamer}
Zhengyi Wang, Cheng Lu, Yikai Wang, Fan Bao, Chongxuan Li, Hang Su, and Jun Zhu.
\newblock Prolificdreamer: High-fidelity and diverse text-to-3d generation with variational score distillation.
\newblock \emph{NeurIPS}, 2023{\natexlab{b}}.

\bibitem[Wang et~al.(2025)Wang, Wang, Chen, Xiang, Chen, Yu, Li, Su, and Zhu]{wang2025crm}
Zhengyi Wang, Yikai Wang, Yifei Chen, Chendong Xiang, Shuo Chen, Dajiang Yu, Chongxuan Li, Hang Su, and Jun Zhu.
\newblock Crm: Single image to 3d textured mesh with convolutional reconstruction model.
\newblock In \emph{European Conference on Computer Vision}, pages 57--74. Springer, 2025.

\bibitem[Weng et~al.(2023)Weng, Yang, Wang, Li, Zhang, Chen, and Zhang]{consistent123}
Haohan Weng, Tianyu Yang, Jianan Wang, Yu Li, Tong Zhang, CL Chen, and Lei Zhang.
\newblock Consistent123: Improve consistency for one image to 3d object synthesis.
\newblock \emph{arXiv preprint arXiv:2310.08092}, 2023.

\bibitem[Wu et~al.(2024)Wu, Yang, Li, Zhang, Liu, Guibas, Lin, and Wetzstein]{wu2023gpteval3d}
Tong Wu, Guandao Yang, Zhibing Li, Kai Zhang, Ziwei Liu, Leonidas Guibas, Dahua Lin, and Gordon Wetzstein.
\newblock Gpt-4v(ision) is a human-aligned evaluator for text-to-3d generation.
\newblock In \emph{CVPR}, 2024.

\bibitem[Xie et~al.(2024)Xie, Lorraine, Cao, Gao, Lucas, Torralba, Fidler, and Zeng]{xie2024latte3d}
Kevin Xie, Jonathan Lorraine, Tianshi Cao, Jun Gao, James Lucas, Antonio Torralba, Sanja Fidler, and Xiaohui Zeng.
\newblock Latte3d: Large-scale amortized text-to-enhanced3d synthesis.
\newblock \emph{The 18th European Conference on Computer Vision (ECCV)}, 2024.

\bibitem[Xu et~al.(2023)Xu, Liu, Wu, Tong, Li, Ding, Tang, and Dong]{image_reward}
Jiazheng Xu, Xiao Liu, Yuchen Wu, Yuxuan Tong, Qinkai Li, Ming Ding, Jie Tang, and Yuxiao Dong.
\newblock Imagereward: Learning and evaluating human preferences for text-to-image generation.
\newblock In \emph{Advances in Neural Information Processing Systems}, pages 15903--15935. Curran Associates, Inc., 2023.

\bibitem[Xu et~al.(2024)Xu, Cheng, Gao, Wang, Gao, and Shan]{xu2024instantmesh}
Jiale Xu, Weihao Cheng, Yiming Gao, Xintao Wang, Shenghua Gao, and Ying Shan.
\newblock Instantmesh: Efficient 3d mesh generation from a single image with sparse-view large reconstruction models.
\newblock \emph{arXiv preprint arXiv:2404.07191}, 2024.

\bibitem[Yan et~al.(2024)Yan, Wu, and Ma]{yan2024flow}
Runjie Yan, Kailu Wu, and Kaisheng Ma.
\newblock Flow score distillation for diverse text-to-3d generation.
\newblock \emph{arXiv preprint arXiv:2405.10988}, 2024.

\bibitem[Ye et~al.(2023)Ye, Zhang, Liu, Han, and Yang]{ipadapter}
Hu Ye, Jun Zhang, Sibo Liu, Xiao Han, and Wei Yang.
\newblock Ip-adapter: Text compatible image prompt adapter for text-to-image diffusion models.
\newblock \emph{arXiv preprint arXiv:2308.06721}, 2023.

\bibitem[Yi et~al.(2024)Yi, Fang, Wang, Wu, Xie, Zhang, Liu, Tian, and Wang]{yi2023gaussiandreamer}
Taoran Yi, Jiemin Fang, Junjie Wang, Guanjun Wu, Lingxi Xie, Xiaopeng Zhang, Wenyu Liu, Qi Tian, and Xinggang Wang.
\newblock Gaussiandreamer: Fast generation from text to 3d gaussians by bridging 2d and 3d diffusion models.
\newblock In \emph{CVPR}, 2024.

\bibitem[Yu et~al.(2022)Yu, Xu, Koh, Luong, Baid, Wang, Vasudevan, Ku, Yang, Ayan, Hutchinson, Han, Parekh, Li, Zhang, Baldridge, and Wu]{yu2022scaling}
Jiahui Yu, Yuanzhong Xu, Jing~Yu Koh, Thang Luong, Gunjan Baid, Zirui Wang, Vijay Vasudevan, Alexander Ku, Yinfei Yang, Burcu~Karagol Ayan, Ben Hutchinson, Wei Han, Zarana Parekh, Xin Li, Han Zhang, Jason Baldridge, and Yonghui Wu.
\newblock Scaling autoregressive models for content-rich text-to-image generation.
\newblock \emph{Transactions on Machine Learning Research}, 2022.

\bibitem[Yu et~al.(2024{\natexlab{a}})Yu, Guo, Li, Liang, Zhang, and Qi]{CSD}
Xin Yu, Yuan-Chen Guo, Yangguang Li, Ding Liang, Song-Hai Zhang, and Xiaojuan Qi.
\newblock Text-to-3d with classifier score distillation.
\newblock In \emph{The Twelfth International Conference on Learning Representations}, 2024{\natexlab{a}}.

\bibitem[Yu et~al.(2024{\natexlab{b}})Yu, Guo, Li, Liang, Zhang, and Qi]{yu2024texttod}
Xin Yu, Yuan-Chen Guo, Yangguang Li, Ding Liang, Song-Hai Zhang, and Xiaojuan Qi.
\newblock Text-to-3d with classifier score distillation.
\newblock In \emph{The Twelfth International Conference on Learning Representations}, 2024{\natexlab{b}}.

\bibitem[Zhang et~al.(2024)Zhang, Bi, Tan, Xiangli, Zhao, Sunkavalli, and Xu]{gslrm2024}
Kai Zhang, Sai Bi, Hao Tan, Yuanbo Xiangli, Nanxuan Zhao, Kalyan Sunkavalli, and Zexiang Xu.
\newblock Gs-lrm: Large reconstruction model for 3d gaussian splatting.
\newblock \emph{European Conference on Computer Vision}, 2024.

\end{thebibliography}
}
\clearpage
\setcounter{page}{1}
\maketitlesupplementary

\setcounter{section}{0}

In this supplementary material, we provide additional evaluations of our method. 
\cref{sec:additional_quantitative} includes a quantitative assessment using GPTEval3D \cite{wu2023gpteval3d}, a human-aligned evaluator for text-to-3D generation. 
\cref{sec:addtional_qualitative} presents more qualitative results of the generated 3D models including results with 3DGS, diverse results with the same prompts provided different reference images, and more qualitative results. 

\section{Evaluation with GPTEval3D}
\label{sec:additional_quantitative}
We provide an additional assessment of our methods using GPTEval3D \cite{wu2023gpteval3d}, a human-aligned evaluator for text-to-3D generation. 
Specifically, GPTEval3D evaluates the quality of 3D objects over 110 prompts across five criteria: Text-Asset Alignment, 3D Plausibility, Text-Geometry Alignment, Texture Details, and Geometry Details. For each evaluation, a pair of objects generated by two different methods, along with an input prompt and an evaluation criterion, is submitted to a large language model (e.g., GPT), which outputs its preference for one object over the other. This process is analogous to a user study but conducted by a large language model. The scores are aggregated to calculate an Elo score, with higher scores indicating better performance in the respective criteria~\cite{wu2023gpteval3d}. 

The quantitative results are presented in \cref{tab:gpt_eval3d}. Our methods achieve the highest overall score and maintain superior performance in four out of five criteria, except Text-Asset Alignment, where our method is ranked second to RichDreamer \cite{qiu2024richdreamer}. These quantitative results, along with those provided by T3Bench \cite{t3bench} (Tab.~1 main paper), demonstrate that our method can generate high-quality 3D objects with strong text alignment.

\begin{table*}[t]
    \centering
    \setlength{\tabcolsep}{5pt}
    \begin{tabular}{lp{2cm}p{1.6cm}p{2.5cm}p{1.4cm}p{1.6cm}p{2cm}}
        \toprule
        \textbf{Methods} & \textbf{Text-Asset Alignment} $\uparrow$ & \textbf{3D  Plausibility} $\uparrow$ & \textbf{Text-Geometry Alignment} $\uparrow$ & \textbf{Texture Details} $\uparrow$ & \textbf{Geometry Details} $\uparrow$ & \textbf{Overall} $\uparrow$ \\
        \midrule
        RichDreamer \cite{qiu2024richdreamer}      & \textbf{1295} & \underline{1225} & \underline{1260} & \underline{1356} & 1251 & \underline{1277} \\
        MVDream \cite{shi2024mvdream}          & 1271 & 1147 & 1251 & 1325 & \underline{1255} & 1250 \\
        ProlificDreamer \cite{wang2023prolificdreamer}  & 1262 & 1059 & 1152 & 1246 & 1181 & 1180 \\
        LatentNeRF \cite{Metzer_2023_CVPR}       & 1222 & 1145 & 1157 & 1180 & 1161 & 1173 \\
        Instant3D \cite{li2024instantd}        & 1200 & 1088 & 1153 & 1152 & 1181 & 1155 \\
        Magic3D \cite{magic3d}          & 1152 & 1001 & 1084 & 1178 & 1100 & 1100 \\
        DreamGaussian \cite{tang2024dreamgaussian}    & 1101 & 954 & 1159 & 1126 & 1131 & 1094 \\
        SJC \cite{sjc}              & 1130 & 995 & 1034 & 1080 & 1043 & 1056 \\
        Fantasia3D \cite{fantasia3d}       & 1068 & 892 & 1006 & 1109 & 1027 & 1021 \\
        Dreamfusion \cite{poole2023dreamfusion}      & 1000 & 1000 & 1000 & 1000 & 1000 & 1000 \\
        One2345 \cite{one2345}          & 872 & 829 & 850 & 911 & 860 & 864 \\
        Shap-E \cite{shap-e}         & 843 & 842 & 846 & 784 & 846 & 836 \\
        Point-E \cite{point-e}         & 725 & 690 & 689 & 716 & 746 & 713 \\
        \midrule
        ISD (ours)              & \underline{1291} & \textbf{1271} & \textbf{1269} & \textbf{1370} & \textbf{1266} & \textbf{1294} \\
        \bottomrule
    \end{tabular}
    \caption{Comparision with text-to-3D methods using GPTEval3D \cite{wu2023gpteval3d} benchmark. The best results are bold while the second best results are underlined.}
    \label{tab:gpt_eval3d}
\end{table*}

\section{More Qualitative Results}
\label{sec:addtional_qualitative}

\begin{figure}[h!]
  \centering
  \includegraphics[width=0.9\linewidth]{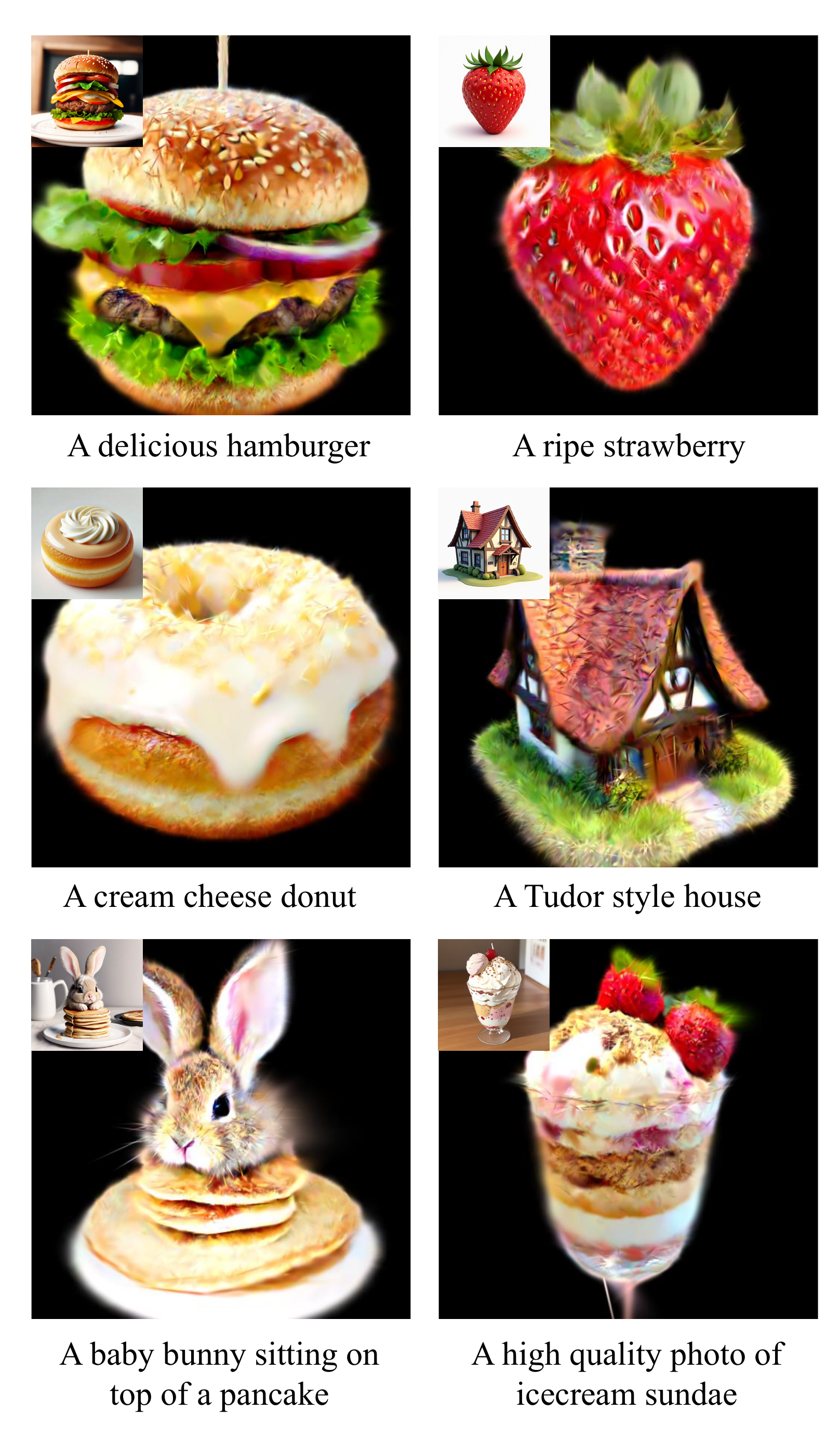}
   \caption{Qualitative results with 3D Gaussian splatting.}
   \label{fig:3dgs}
\end{figure}

\myheading{Qualitative results with 3DGS.}
We further demonstrate the effectiveness of our method by optimizing a 3D Gaussian splatting (3DGS) representation~\cite{kerbl3Dgaussians}.
Our results are shown in \cref{fig:3dgs}, demonstrating that our method generalizes effectively to different 3D representations, including NeRFs and 3DGS.

\begin{figure*}[t]
  \centering
  \includegraphics[width=1.0\linewidth]{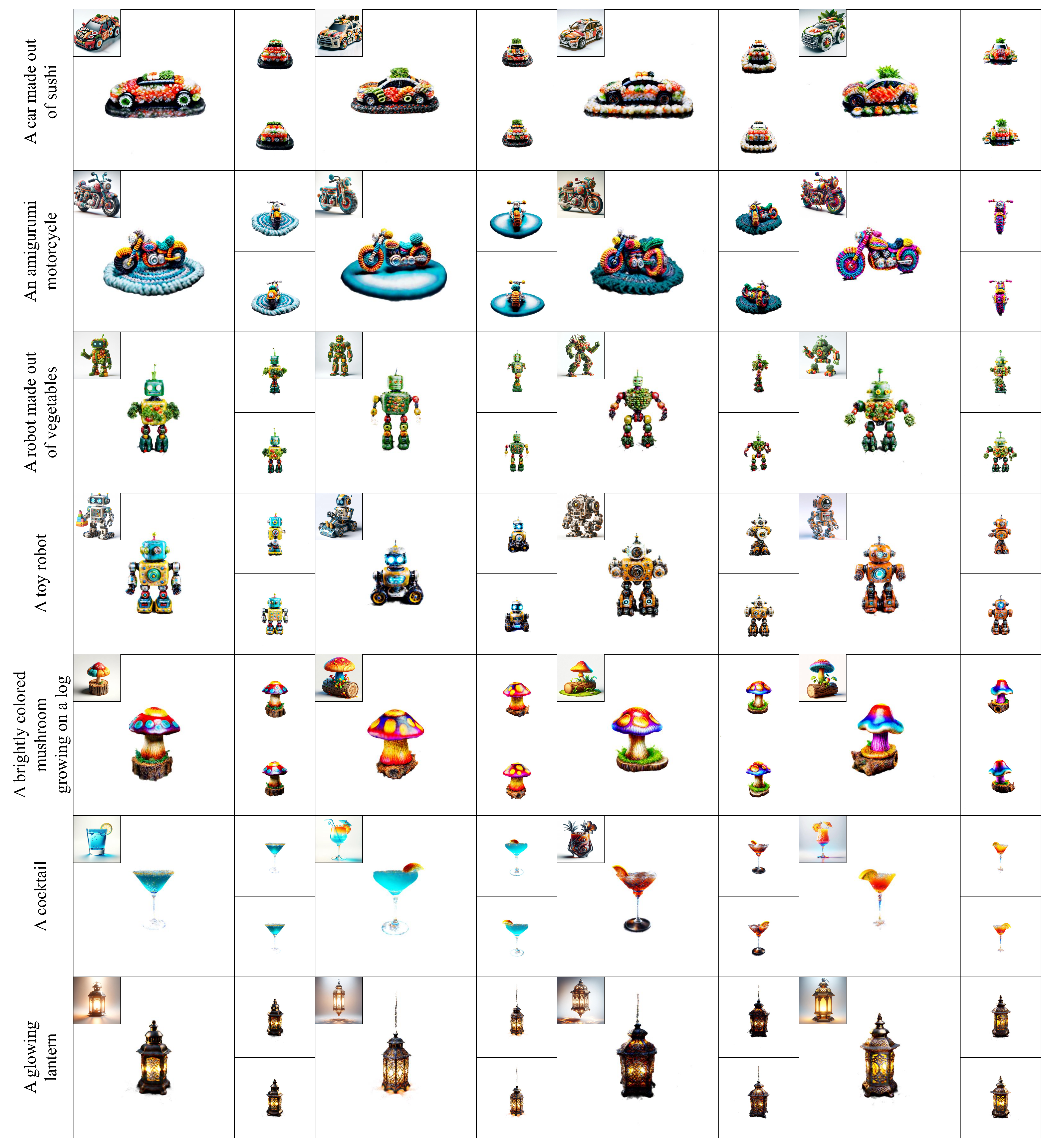}
   \caption{Diverse results of our method when varying the input reference images for each text prompt.}
   \label{fig:diversity}
\end{figure*}

\begin{figure*}[t]
  \centering
  \includegraphics[width=1.0\linewidth]{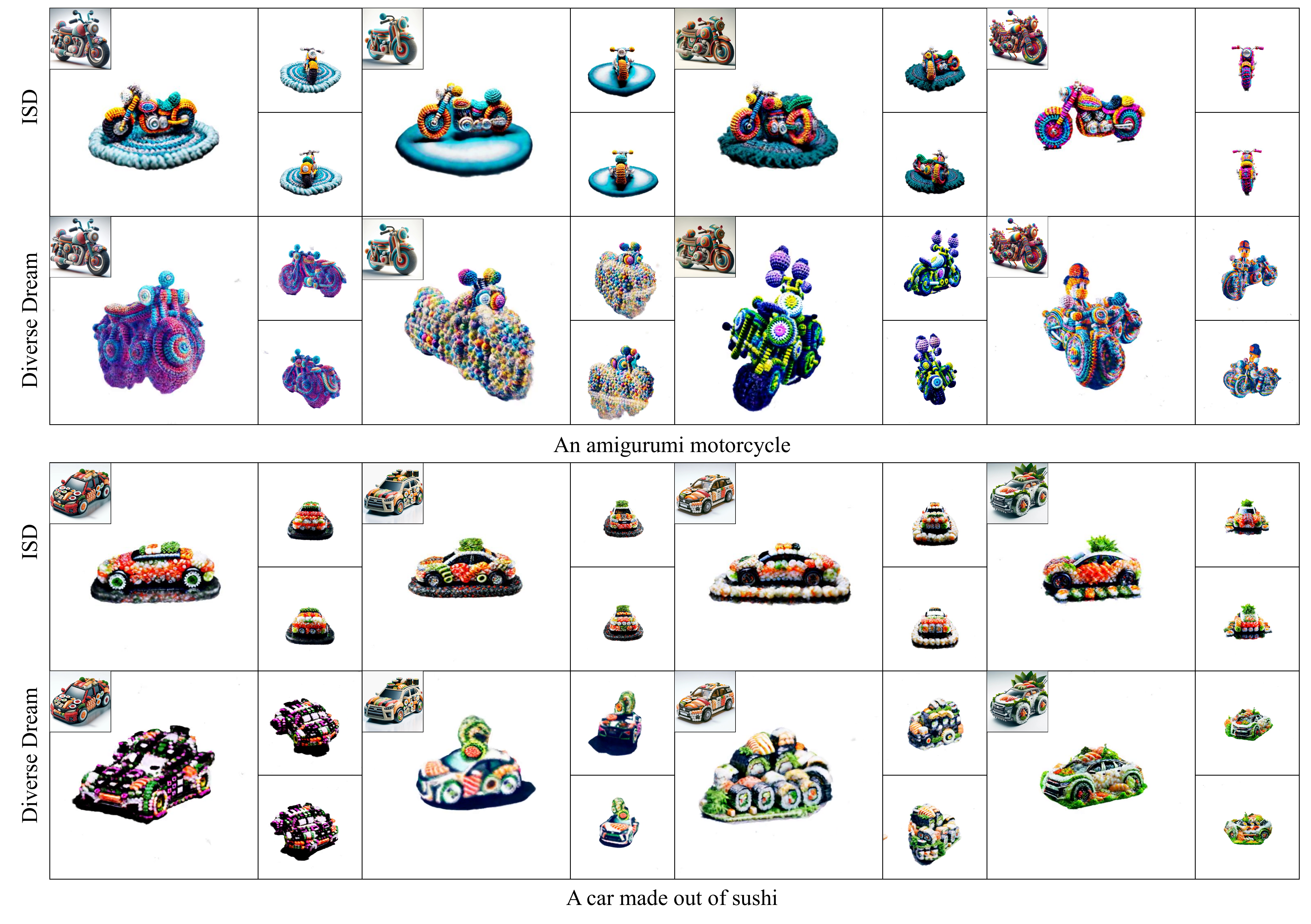}
   \caption{Diverse results of our method when comparing with DiverseDream \cite{tran2025diverse}}
   \label{fig:diversity-vs-diversedream}
\end{figure*}

\myheading{Diversity results.}
We further evaluate the diversity of our 3D generation by considering various reference images sampled with the same prompts.
As shown in \cref{fig:diversity}, each reference image leads to distinct 3D results. Another approach that enhances the diversity of the text-to-3D task is DiverseDream \cite{tran2025diverse}, which utilizes the structure of reference images by employing textual inversion to diversify the generated 3D objects. However, this method is unstable and struggles to generate asymmetric 3D objects, primarily due to overfitting issues associated with the textual inversion stage. In contrast, our proposed method is able to generate high-quality and diverse 3D objects while preserving reasonable geometric integrity. Furthermore, our results better align with the input references compared to DiverseDream \cite{tran2025diverse}, as illustrated in \cref{fig:diversity-vs-diversedream}.

\begin{figure*}[t]
  \centering
  \includegraphics[width=1.0\linewidth]{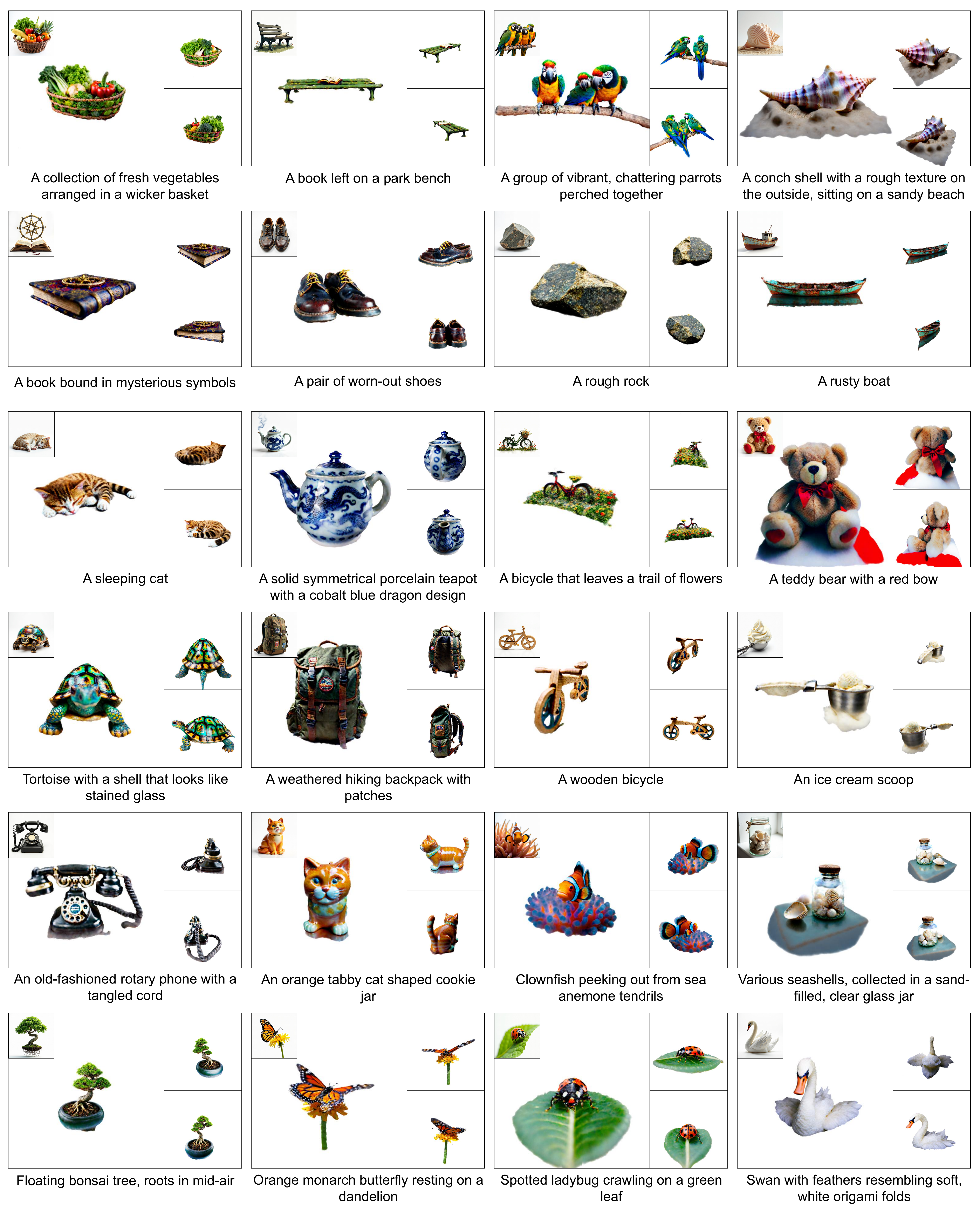}
   \caption{Additional qualitative results (1/3).}
   \label{fig:qualitative_sup_1}
\end{figure*}

\begin{figure*}[t]
  \centering
  \includegraphics[width=1.0\linewidth]{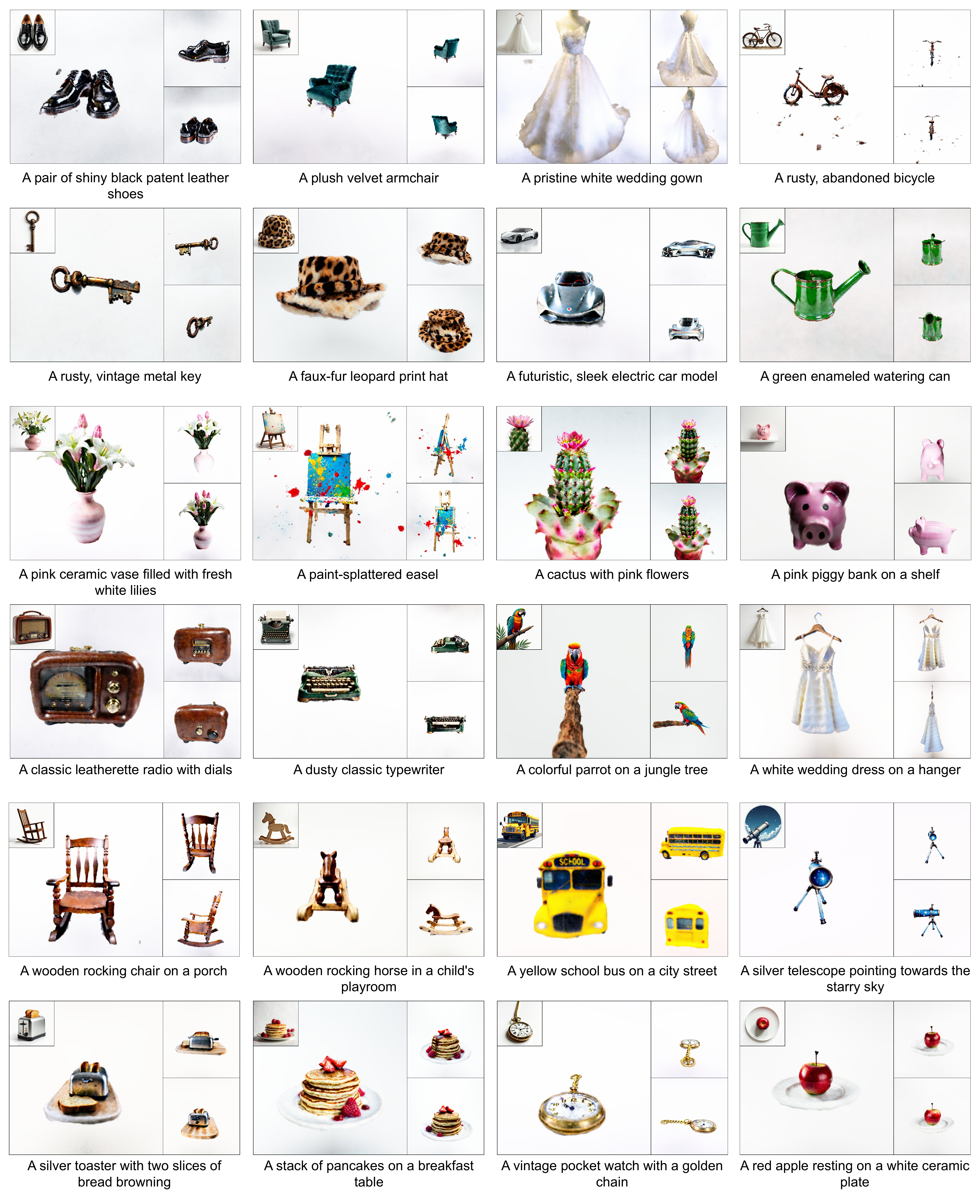}
   \caption{Additional qualitative results (2/3).}
   \label{fig:qualitative_sup_2}
\end{figure*}

\begin{figure*}[t]
  \centering
  \includegraphics[width=1.0\linewidth]{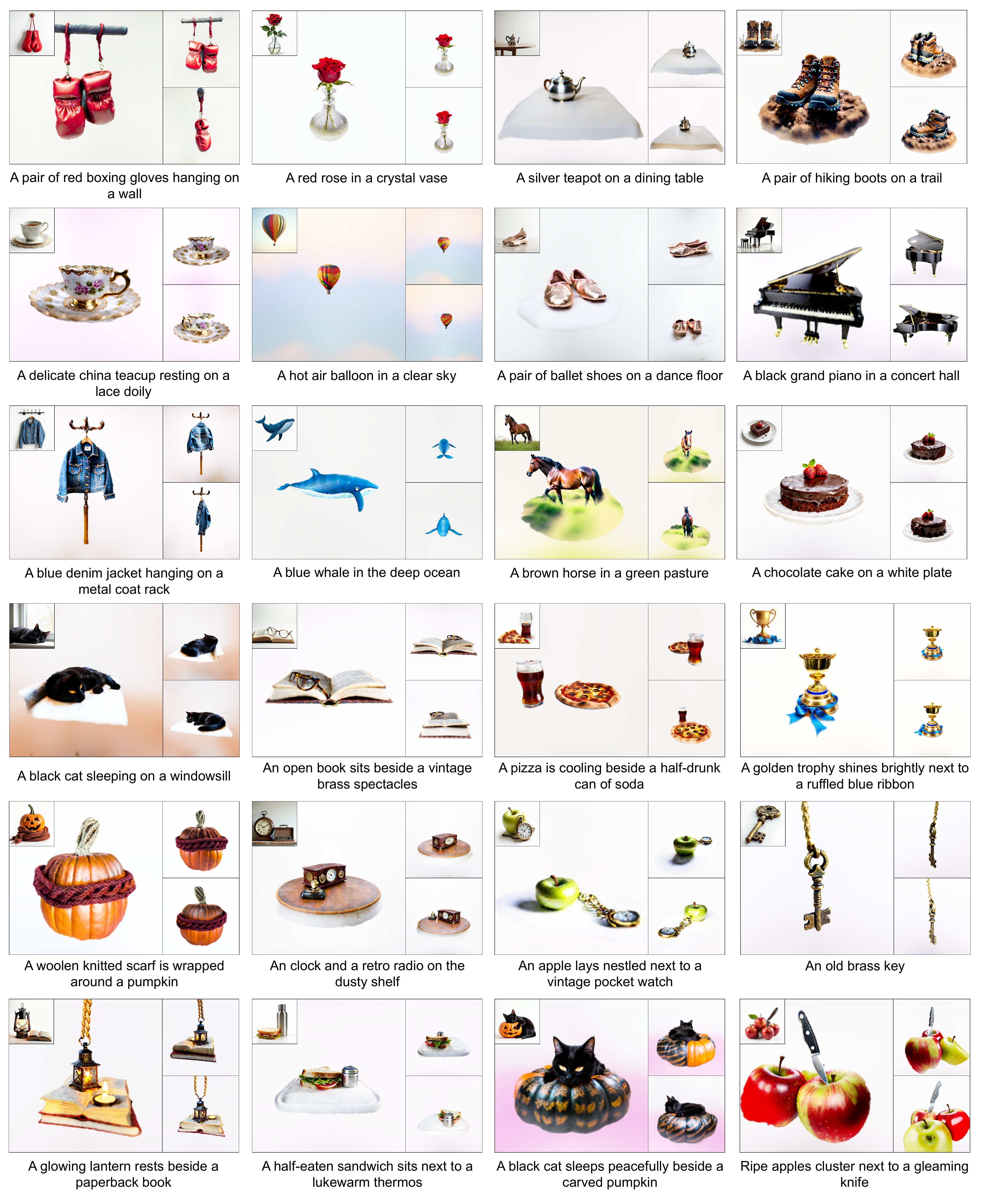}
   \caption{Additional qualitative results (3/3).}
   \label{fig:qualitative_sup_3}
\end{figure*}

\myheading{Additional qualitative results.}
Finally, we present additional qualitative results in \cref{fig:qualitative_sup_1}, \cref{fig:qualitative_sup_2} and \cref{fig:qualitative_sup_3}.
Specifically, we use prompts from two benchmarks, T3Bench and GPTEval3D, along with additional prompts sourced from DreamFusion \cite{poole2023dreamfusion}.

\end{document}